\documentclass[lettersize,journal]{IEEEtran}
\usepackage{amsmath,amsfonts}
\usepackage{algorithmic}
\usepackage{algorithm}
\usepackage{array}
\usepackage[caption=true,font=footnotesize,labelfont=sf,textfont=sf]{subfig}
\usepackage{textcomp}
\usepackage{stfloats}
\usepackage{url}
\usepackage{verbatim}
\usepackage{graphicx}
\usepackage{cite}
\usepackage{amsthm}
\newtheorem{definition}{Definition}
\newtheorem{proposition}{Proposition}
\newtheorem{remark}{Remark}
\newcommand*{\dif}{\mathop{}\!\mathrm{d}}
\usepackage{mathrsfs}
\usepackage{booktabs}
\usepackage{multirow}
\usepackage{xcolor}
\usepackage{dsfont}
\usepackage{float}
\usepackage{bm}

\usepackage{makecell}
\usepackage{hyperref}

\begin{document}

\title{Robust Multivariate Time Series Forecasting against Intra- and Inter-Series Transitional Shift}


\author{Hui~He,
        Qi~Zhang,
        Kun~Yi,
        Xiaojun~Xue,
        Shoujin~Wang,
        Liang~Hu,
        Longbing~Cao,~\IEEEmembership{Senior~Member,~IEEE}
\IEEEcompsocitemizethanks{\IEEEcompsocthanksitem Hui He is with the School of Medical Technology, Beijing Institute of Technology, Beijing 100081, China (e-mail: hehui617@bit.edu.cn).
\IEEEcompsocthanksitem Qi Zhang and Liang Hu are with the Department of Computer Science, Tongji University, Shanghai 201804, China (e-mail: \{zhangqi\_cs, lianghu\}@tongji.edu.cn).
\IEEEcompsocthanksitem Kun Yi and Xiaojun Xue are with the School of Computer Science and Technology, Beijing Institute of Technology, Beijing 100081, China (e-mail: \{yikun, xiaojunx\}@bit.edu.cn).
\IEEEcompsocthanksitem Shoujin Wang is with the Data Science Institute, University of Technology Sydney, Ultimo, NSW 2007, Australia (e-mail: shoujin.wang@uts.edu.au).
\IEEEcompsocthanksitem Longbing Cao is with the School of Computing, Macquarie University, Sydney, NSW 2109, Australia (e-mail: longbing.cao@mq.edu.au).
}
\thanks{Manuscript received April 19, 2021; revised August 16, 2021.}}

\markboth{Journal of \LaTeX\ Class Files,~Vol.~14, No.~8, August~2021}%
{Shell \MakeLowercase{\textit{et al.}}: A Sample Article Using IEEEtran.cls for IEEE Journals}


\maketitle

\begin{abstract}
The non-stationary nature of real-world Multivariate Time Series (MTS) data presents forecasting models with a formidable challenge of the time-variant distribution of time series, referred to as \textit{distribution shift}. Existing studies on the distribution shift mostly adhere to adaptive normalization techniques for alleviating temporal mean and covariance shifts or time-variant modeling for capturing temporal shifts. Despite improving model generalization, these normalization-based methods often assume a time-invariant transition between outputs and inputs but disregard specific intra-/inter-series correlations, while time-variant models overlook the intrinsic causes of the distribution shift. This limits model expressiveness and interpretability of tackling the distribution shift for MTS forecasting. To mitigate such a dilemma, we present a unified \textit{P}robabilistic \textit{G}raphical \textit{M}odel to \textit{Joint}ly capturing intra-/inter-series correlations and modeling the time-variant transitional distribution, and instantiate a neural framework called \textit{JointPGM} for non-stationary MTS forecasting. Specifically, \textit{JointPGM} first employs multiple Fourier basis functions to learn dynamic time factors and designs two distinct learners: intra-series and inter-series learners. The intra-series learner effectively captures temporal dynamics by utilizing temporal gates, while the inter-series learner explicitly models spatial dynamics through multi-hop propagation, incorporating Gumbel-softmax sampling. These two types of series dynamics are subsequently fused into a latent variable, which is inversely employed to infer time factors, generate final prediction, and perform reconstruction. We validate the effectiveness and efficiency of \textit{JointPGM} through extensive experiments on six highly non-stationary MTS datasets, achieving state-of-the-art forecasting performance of MTS forecasting.
\end{abstract}


\section{Introduction} \label{sec1}
\IEEEPARstart{M}{ultivariate} Time Series (MTS) forecasting has been playing an increasingly ubiquitous role in real-world applications, such as weather condition estimation~\cite{DBLP:conf/aaai/HanLZXD21}, stock trend analysis~\cite{DBLP:conf/kdd/ZhaoKS23,10462534}, electricity consumption planning~\cite{DBLP:conf/nips/CaoWDZZHTXBTZ20}, and traffic flow and speed prediction~\cite{DBLP:conf/ijcai/YuYZ18,DBLP:conf/icde/0001LGWZSHW23}. 
Impressively, various deep learning-based approaches have emerged and led to a surge in deep MTS forecasting models. These approaches elaborately capture complex temporal variations by Temporal Convolution Networks (TCNs)~\cite{DBLP:conf/nips/SenYD19,DBLP:conf/iclr/WuHLZ0L23,DBLP:conf/nips/LiuZCXLM022}, Recurrent Neural Networks (RNNs)~\cite{DBLP:conf/sigir/LaiCYL18,DBLP:journals/corr/FlunkertSG17}, and Transformers~\cite{DBLP:conf/aaai/ZhouZPZLXZ21, DBLP:conf/nips/WuXWL21,DBLP:conf/iclr/NieNSK23,DBLP:conf/icml/ZhouMWW0022}, or explore specific variable-wise 
dependencies by Graph Neural Networks (GNNs)~\cite{DBLP:conf/ijcai/WuPLJZ19,DBLP:conf/nips/0001YL0020,DBLP:conf/kdd/WuPL0CZ20,DBLP:conf/aaai/YangLWZPW23,DBLP:conf/nips/YiZFHHWACN23}. 
Despite the remarkable performance, they fall short in adapting to real-world scenarios where the distributions (\textit{a.k.a.}, statistical properties) of time series change over time due to dynamic generation mechanisms. This phenomenon, known as \textit{distribution shift}~\cite{DBLP:conf/aaai/0010WWWZF23, DBLP:conf/nips/LiuWWL22,10509830}, exposes time series' \textit{highly dynamic} and \textit{non-stationary} nature. It poses significant challenges for forecasting models in effective generalization to varying distributions. Such vulnerability to rapid distributional changes~\cite{DBLP:journals/pami/ZhaoCL23} ultimately results in a dramatic decline in forecasting accuracy over time~\cite{10504612}.



Researchers have explored two primary categories of approaches to tackle the distribution shift in MTS forecasting. 
The first category involves normalization techniques to align time series instances based on the Gaussian assumption. They normalize the input and denormalize the output using adaptively learned statistics (\textit{e.g.}, mean and variance)~\cite{DBLP:conf/cikm/Du0FPQXW21,DBLP:conf/iclr/KimKTPCC22,DBLP:conf/nips/LiuWWL22,DBLP:conf/aaai/0010WWWZF23,DBLP:conf/nips/LiuCLHLXC23} to alleviate the temporal \textit{mean and covariance shift} among instances or between inputs and outputs. Most of these methods, however, assume a \textit{time-invariant transitional distribution} between \textit{output predictions} and \textit{input observations}, \textit{i.e.}, $\mathcal{P}(\bm{x}_{u:u+H}|\bm{x}_{u-L:u})=\mathcal{P}(\bm{x}_{v:v+H}|\bm{x}_{v-L:v})$ at any two steps $u\neq v$. This assumption severely simplifies the non-stationarity of time series and is not consistent with the practical distribution shift~\cite{DBLP:journals/pami/LiCFHZ24,DBLP:conf/iclr/PodkopaevR22}. 
Give an easily understandable example: In stock prediction, the financial factors $\mathcal{P}(\bm{x}_t)$ naturally change due to market fluctuations. Meanwhile, economic laws $\mathcal{P}(\bm{x}_{t:t+h}|\bm{x}_{t-L:t})$ are also vulnerable to abrupt policies, such as government price controls. Additionally, these methods focus on exploring the variable-wise data distribution, overlooking specific intra-/inter-series correlations~\cite{DBLP:journals/expert/Cao22h,DBLP:journals/tois/YiZHSHAN24}. As a result, they struggle to effectively address the counterpart non-stationarity, especially the distribution shift along with inter-series dynamics.

The second category aims to model \textit{time-variant transitional distribution}, \textit{i.e.,} $\mathcal{P}_t(\bm{x}_{t:t+H}|\bm{x}_{t-L:t})$ adapted to any time step $t$, to improve models' temporal generalization. Many advanced models within this category have integrated time information to enhance forecasting performance, indicating that time information enables the models to effectively capture the time-variant characteristics of time series, thereby alleviating the issue of non-stationarity~\cite{DBLP:conf/aaai/ZhouZPZLXZ21,DBLP:conf/nips/WuXWL21,DBLP:conf/icml/ZhouMWW0022}.
Herein, some models incorporate temporal meta-knowledge to correct the bias caused by the distribution shift within a discriminative meta-learning framework~\cite{DBLP:conf/cikm/YouZDFH21,DBLP:conf/aaai/LiYLX022,DBLP:conf/kdd/ZhaoKS23}. Another representative study~\cite{DBLP:conf/nips/LiuLWL23} utilizes Koopman operators as linear portraits of implicit transitions to approximate time-invariant and time-variant dynamics. However, these methods often overlook the underlying causes of the distribution shift, which results in a prediction/generation process that resembles a black box. Consequently, the models' expressiveness and interoperability, both crucial in various real-world applications, become limited~\cite{10457027}. For instance, the distribution of traffic flow on a road can undergo sudden changes triggered by various events, such as unexpected temperature spikes on that road or traffic accidents on connected routes.



Upon a deeper examination of MTS forecasting, it is evident that a prevalent and advanced approach to capture intra-/inter series correlations involves decomposing the transitional distribution into intra-/inter-series transitional distributions, \textit{i.e.}, $\mathcal{P}_t(\bm{x}_{t:t+H}^{(i)}|\bm{X}_{t-L:t})=\mathcal{P}_t(\bm{x}_{t:t+H}^{(i)}|\bm{x}_{t-L:t}^{(i)})\mathcal{P}_t(\bm{x}_{t:t+H}^{(i)}|\bm{X}_{t-L:t}^{(\overline{i})})$, where $\bm{X}^{(\overline{i})}$ refers to all the variables excluding the $i^{th}$ variable.
Consequently, we argue that the observed non-stationarity in MTS is primarily attributed to the implicitly time-variant intra- and inter-series correlations~\cite{DBLP:conf/aaai/CaiLLFW24,DBLP:conf/nips/ZhangWWCSZWJT23}. 
Based on this insight, we decompose the transitional shift into intra-series transitional shift, \textit{i.e.}, $\mathcal{P}(\bm{x}_{u:u+H}^{(i)}|\bm{x}_{u-L:u}^{(i)}) \neq \mathcal{P}(\bm{x}_{v:v+H}^{(i)}|\bm{x}_{v-L:v}^{(i)})$, and the inter-series transitional shift, \textit{i.e.}, $\mathcal{P}(\bm{x}_{u:u+H}^{(i)}|\bm{x}_{u-L:u}^{(\overline{i})}) \neq \mathcal{P}(\bm{x}_{v:v+H}^{(i)}|\bm{x}_{v-L:v}^{(\overline{i})})$.
Compared to quantifying the superficial data distribution, \textit{i.e.}, $\mathcal{P}(\bm{x}_t)$, exploring the transitional distribution is more rational and practical. Estimating data distributions can be more challenging, whereas conditional distributions are easily available and offer a more intuitive understanding of the generation of MTS data.
Furthermore, compared to previous time-variant models with coarse-grained transitional distribution, this decomposition corresponds to a general modeling approach that learns observation-prediction projections and accounts for intra- and inter-series correlations. It reveals the intrinsic causes of the transitional distribution, ensuring desirable model interpretability and the potential to enhance forecasting performance by jointly addressing the distribution shift and modeling intra-/inter-series correlations.

In light of the above discussion, we devise a unified \textit{P}robabilistic \textit{G}raphical \textit{M}odel which delves to \textit{Joint}ly capturing intra-/inter-series correlations and modeling the time-variant transitional distribution, and instantiate a neural framework, named \textit{JointPGM} for non-stationary MTS forecasting. 
Known for effectively representing complex distribution and statistical relationships among variables in a principled and interpretable manner, PGM is an underexplored but promising framework for robust MTS forecasting against distribution shifts.
We organize \textit{JointPGM} as a dual-encoder architecture, which includes a time factor encoder (TFE) and an independence-based series encoder (ISE). Technically, in TFE, \textit{JointPGM} employs multiple Fourier basis functions to capture dynamic time factors and introduces linear projections to learn the mean and variance of Gaussian sampling. Corresponding to the intra-/inter-series transitional shifts, \textit{JointPGM} sequentially employs two distinct learners in ISE: intra-series and inter-series learners. The intra-series learner focuses on capturing temporal dynamics within each series and utilizes a temporal gate adjusted by learned time factors to control the message passing of temporal features, making them sensitive to non-stationary environments. The inter-series learner explicitly models spatial dynamics through multi-hop propagation incorporating Gumbel-softmax sampling. These two types of series dynamics are then fused and transformed into the latent variable, which is inversely used to infer time factors, generate final prediction, and perform reconstruction. 
We incorporate various constraints on all the above sub-processes based on a tailored PGM framework with theoretical guarantees, ensuring a clear understanding of the role played by each sub-process in forecasting MTS with non-stationarity.

To summarize, our key contributions are as follows:
\begin{itemize}
\item Going beyond previous methods, we propose \textit{JointPGM}, an effective neural framework for MTS forecasting under fine-grained transitional shift, built upon a probabilistic graphical model with jointly addressing the distribution shift and modeling intra-/inter-series correlations.


\item To achieve \textit{JointPGM}, we elaborately design two distinct learners: an intra-series learner to capture temporal dynamics via temporal gates and an inter-series learner to explicitly model spatial dynamics through multi-hop propagation by incorporating Gumbel-softmax sampling.
\item We conduct extensive experiments on six highly non-stationary MTS datasets, achieving state-of-the-art performance with an average improvement of 15.3\% in MAE and 37.9\% in MSE over all baselines for forecasting. 
\end{itemize}

\section{Related Work}  \label{sec2}

\subsection{Deep Models for Multivariate Time Series Forecasting}
Multivariate time series (MTS) forecasting is a longstanding research topic~\cite{yi2023surveydeeplearningbased,DBLP:journals/ijon/BaiZHHWN23}. 
Initially, traditional statistical models such as Gaussian process (GP)~\cite{DBLP:conf/nips/SalinasBCMG19} have been proposed for their appealing simplicity and interpretability. 
Recently, with the bloom of deep learning, many deep models with elaboratively designed architectures have made great breakthroughs in capturing intra- and inter-series correlations for MTS forecasting. On one hand, the RNN-~\cite{DBLP:conf/sigir/LaiCYL18,DBLP:journals/corr/FlunkertSG17} and TCN-based~\cite{DBLP:conf/nips/SenYD19,DBLP:conf/iclr/WuHLZ0L23,DBLP:conf/nips/LiuZCXLM022} models have shown competitiveness in modeling complex temporal relationships. However, due to their recurrent structures or the locality property of one-dimensional convolutional kernels, they are limited in handling long-term dependencies. 
Soon afterward, Transformer and its variants~\cite{DBLP:conf/aaai/ZhouZPZLXZ21,DBLP:conf/nips/WuXWL21,DBLP:conf/icml/ZhouMWW0022,DBLP:conf/iclr/NieNSK23,DBLP:journals/tkde/HeZWYNC24} have achieved superior performance on MTS forecasting, particularly notable in scenarios with long prediction lengths. They focus on renovating the canonical structure and designing a novel attention mechanism to reduce the quadratic complexity while automatically learning the correlations between elements in a series. 
Despite the complicated design of Transformer-based models, recent MLP-based models~\cite{DBLP:conf/nips/YiZFWWHALCN23,DBLP:conf/aaai/ZengCZ023,DBLP:conf/kdd/EkambaramJNSK23,DBLP:journals/pvldb/ZhongSZLLC24} with simple structure and low complexity can surpass previous models across various common benchmarks for MTS forecasting. 
Another crucial aspect of MTS forecasting involves capturing the correlations among multiple time series. Current models highly depend on GNNs~\cite{DBLP:conf/ijcai/WuPLJZ19,DBLP:conf/nips/0001YL0020,DBLP:conf/kdd/WuPL0CZ20,DBLP:conf/aaai/YangLWZPW23,DBLP:conf/nips/YiZFHHWACN23,DBLP:conf/nips/CaoWDZZHTXBTZ20,DBLP:journals/pvldb/ZhaoGCHZY23} or ordered tree~\cite{DBLP:conf/aaai/HeZBYN22} due to their remarkable capability in modeling structural dependencies. Most of them can automatically learn the topological structure of inter-series correlations by leveraging node similarity~\cite{DBLP:conf/nips/0001YL0020,DBLP:conf/kdd/WuPL0CZ20,DBLP:conf/aaai/YangLWZPW23} or self-attention mechanism~\cite{DBLP:conf/nips/CaoWDZZHTXBTZ20}.
More recently, Crossformer~\cite{DBLP:conf/iclr/ZhangY23} and iTransformer~\cite{liu2024itransformer} have been specifically proposed to explicitly capture the mutual interactions among multiple variables by refurbishing the architecture and components such as the attention module of Transformer. 
Different from previous works focusing on better modeling temporal relationships within and among time series, we analyze the MTS forecasting task from a more fundamental review of the non-stationary nature, which constitutes an indispensable property of MTS data.

\subsection{Improving Robustness against Distribution Shifts}
Despite many remarkable deep models, MTS forecasting still suffers from severe distribution shifts~\cite{fan2024deepfrequencyderivativelearning} considering the distribution of real-world series changes temporally. 
To improve robustness over varying distributions, one category of widely-explored methods~\cite{DBLP:journals/tnn/PassalisTKGI20,DBLP:conf/iclr/KimKTPCC22,DBLP:conf/nips/LiuWWL22,DBLP:conf/aaai/0010WWWZF23,DBLP:conf/cikm/Du0FPQXW21,DBLP:conf/nips/LiuCLHLXC23} stationarize deep model inputs by the normalization techniques.
For example, 
RevIN~\cite{DBLP:conf/iclr/KimKTPCC22} proposes a reversible instance normalization technique to reduce temporal distribution shift. 
Based on RevIN, Dish-TS~\cite{DBLP:conf/aaai/0010WWWZF23} designs a dual coefficient network to learn two sets of distribution coefficients and captures the distribution shift between inputs and outputs.
Stationary~\cite{DBLP:conf/nips/LiuWWL22} adopts de-stationary attention to handle the over-stationarization issue which may damage the model's capability of modeling specific temporal dependency.
SAN~\cite{DBLP:conf/nips/LiuCLHLXC23} utilizes slice-level adaptive normalization to mitigate non-stationarity.
However, these methods typically assume a time-invariant transitional distribution and overlook the distribution shift caused by inter-series dynamics.
Another category~\cite{DBLP:conf/cikm/YouZDFH21,DBLP:conf/aaai/LiYLX022,DBLP:conf/kdd/ZhaoKS23} learns to model time-variant transitional distribution by incorporating temporal meta-knowledge to correct the bias caused by distribution shift in a discriminative meta-learning framework,
which is generally designed for bridging the gap between the training and test data.
More recently, Koopman predictors Koopa~\cite{DBLP:conf/nips/LiuLWL23} and KNF~\cite{DBLP:conf/iclr/WangDAY23} employ Koopman operators as linear portraits of implicit transitions to capture time-invariant and time-variant dynamics.
While these models model the time-variant transitional distributions, such coarse-grained modeling fails to reveal the intrinsic causes of the transitional distribution, limiting the models' interpretability and expressiveness. 
In this paper, we propose \textit{JointPGM} to model the practical transitional distribution and decompose it based on the prevalent approach of learning intra-/inter-series correlations.

\section{Problem Formulations} \label{sec3}
In this section, we start with the formulations of MTS forecasting and define the concepts central to distribution shift. Detailed notations are summarized in Table \ref{table A.1} in Appendix \ref{secA}. 

\textbf{Multivariate Time Series Forecasting.} Let a regularly sampled time series dataset with a total of $N$ distinct time series and $T$ time steps be denoted as $[\bm{x}^{(1)},..., \bm{x}^{(i)},...,\bm{x}^{(N)}] \in \mathbb{R}^{N \times T}$, where $\bm{x}^{(i)} \in \mathbb{R}^T$ denotes the sequence values of time series $i$ at $T$ time steps. Given a lookback window of length-$L$ and a horizon window of length-$H$, the \textit{multivariate time series forecasting} involves utilizing historical multivariate observations $\bm{X}_{t-L:t}=\{\bm{x}_{t-L:t}^{(i)}\}_{i=1}^{N}$ to predict their future multivariate values $\bm{X}_{t:t+H}=\{\bm{x}_{t:t+H}^{(i)}\}_{i=1}^{N}$ at time step $t$. The forecasting process can be formulated as:
\begin{equation}\label{eqn1}
\bm{X}_{t:t+H}=\mathcal{F}_{\Theta}(\bm{X}_{t-L:t})=\mathcal{F}_{\Theta}(\{\bm{x}_{t-L:t}^{(i)}\}_{i=1}^{N})
\end{equation}
where the function map $\mathcal{F}_{\Theta}:\mathbb{R}^{N \times L} \rightarrow \mathbb{R}^{N \times H}$ can be regarded as a forecasting model parameterized by $\Theta$. 


\textbf{Distribution Shift in Time Series.}~Recall the intuitive financial example mentioned in Section \ref{sec1}, where the economic laws are vulnerable to abrupt policy changes. Therefore, we propose to involve the more rational and practical transitional shift assumption and further decompose the \textit{integrated} transitional shift in time series into two types at a finer granularity, namely, \textit{intra-series transitional shift} and \textit{inter-series transitional shift}, with their definitions provided below. 
\begin{definition}[Intra-series Transitional Shift]
    \label{def1}
    Given the $i^{th}$ time series $\bm{x}^{(i)}$, which can be split into several lookback windows $\{\bm{x}_{t-L:t}^{(i)}\}_{t=L}^{T-H}$ and their corresponding horizon windows $\{\bm{x}_{t:t+H}^{(i)}\}_{t=L}^{T-H}$. Intra-series Transitional Shift is referred to the case that the transitional distribution $\mathcal{P}(\bm{x}_{u:u+H}^{(i)}|\bm{x}_{u-L:u}^{(i)}) \neq \mathcal{P}(\bm{x}_{v:v+H}^{(i)}|\bm{x}_{v-L:v}^{(i)})$ for any two time steps $u$ and $v$ with $L \leq u \neq v \leq T-H$.
\end{definition}

\begin{definition}[Inter-series Transitional Shift]
    \label{def2}
    Given the $i^{th}$ time series $\bm{x}^{(i)}$ with its complementary set $\bm{x}^{(\overline{i})}$. Similar to $\bm{x}^{(i)}$, $\bm{x}^{(\overline{i})}$ can be split into several lookback windows $\{\bm{x}_{t-L:t}^{(\overline{i})}\}_{t=L}^{T-H}$ and their corresponding horizon windows $\{\bm{x}_{t:t+H}^{(\overline{i})}\}_{t=L}^{T-H}$. Inter-series Transitional Shift is referred to the case that the transitional distribution $\mathcal{P}(\bm{x}_{u:u+H}^{(i)}|\bm{x}_{u-L:u}^{(\overline{i})}) \neq \mathcal{P}(\bm{x}_{v:v+H}^{(i)}|\bm{x}_{v-L:v}^{(\overline{i})})$ for any two time steps $u$ and $v$ with $L \leq u \neq v \leq T-H$.
\end{definition}

\begin{remark}
{The combination of these two definitions fully describes the complex distribution shifts encountered in reality. The former indicates the variations in transitional distribution for each series, while the latter reflects the variations in transitional distribution among different series. Since characterizing the local relationship between pairwise series in Definition \ref{def2} is overly complex, we describe the relationship between each series and its complementary set from a global perspective.}
\end{remark}

\begin{figure}[!t] 
\centering
\subfloat[
]{
    \label{fig2_1}
    \includegraphics[width=0.315\linewidth]{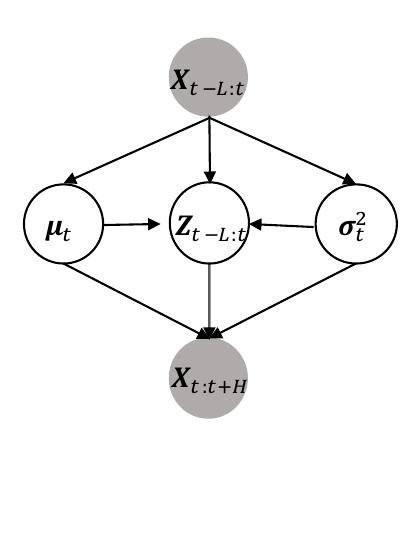}
    }
\subfloat[]{
    \label{fig2_2}
    \includegraphics[width=0.315\linewidth]{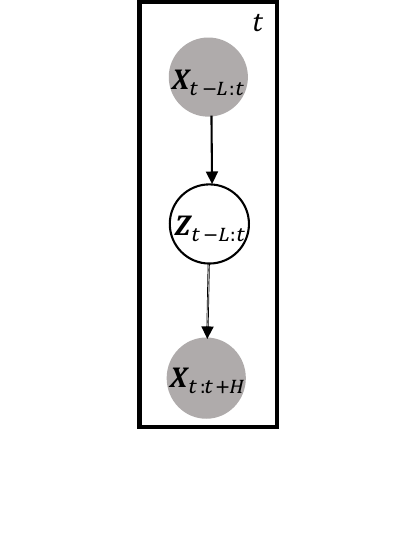}
    }
\subfloat[]{
    \label{fig2_3}
    \includegraphics[width=0.315\linewidth]{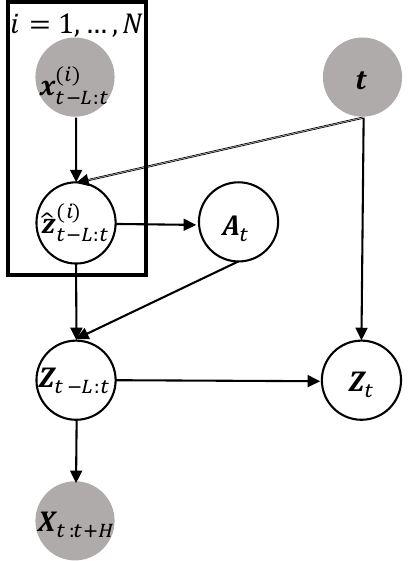}
    }
\caption{Comparison of the graphical representations at time step $t$ for non-stationary MTS forecasting tasks: (a) normalization-based methods, (b) time-variant models, and (c) our proposed \textit{JointPGM}. Grey circles denote observable input-output variables, while white circles denote intermediate-generated latent variables.}
\label{figure 2}
\end{figure}

\begin{figure*}[!t]
  \includegraphics[width=\textwidth]{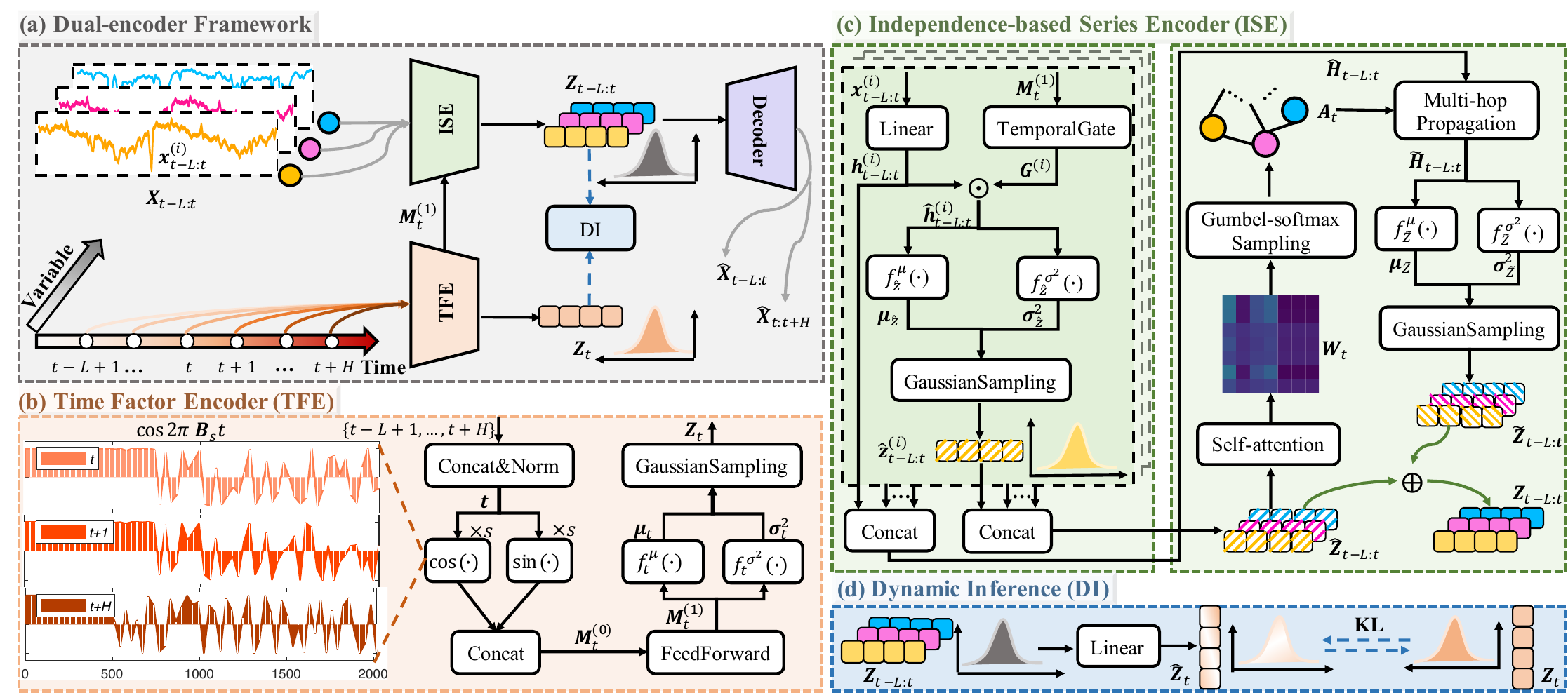}
  \caption{(a) The overview of the proposed \textit{JointPGM} framework featuring dual encoders. Given $N$ time series and $L+H$ time steps as input, \textit{JointPGM} outputs multivariate predictions $\hat{\bm{X}}_{t:t+H}$ and reconstructions $\hat{\bm{X}}_{t-L:t}$. (b) TFE is applied to represent dynamic time factors which can reflect non-stationary environments. (c) ISE captures the intrinsic correlations within each series and among different series using two distinct learners: intra-series learner (left part of ISE) and inter-series learner (right part of ISE), organized in a consecutive fashion. (d) DI aims to reversely infer time factors from time series in latent space, designed to be dynamic to render these factors more discriminative and sensitive to non-stationarity.}
  \label{figure 3}
\end{figure*}

\section{Methodology} 
In this section, we first present our tailored PGM and formally analyze the distinctions between normalization-based methods, time-variant models, and ours in Section \ref{sec3.2}. In Section \ref{sec3.3}, we introduce the corresponding instantiated dual-encoder architecture. Finally, we decompose the learning objective based on the PGM and our purpose in Section \ref{sec3.4}. 

\subsection{Probabilistic Decomposition for Transitional Shift}  \label{sec3.2}
Recall the notations in non-stationary MTS forecasting task, \textit{i.e.}, $\bm{X}_{t-L:t}$, $\bm{X}_{t:t+H}$, and $\bm{t}$, they correspond to the latent variable $\bm{Z}_{t-L:t}$, $\bm{Z}_{t:t+H}$, and $\bm{Z}_t$ respectively. 
We construct the probabilistic graphical model for normalization-based methods, time-variant models, and our proposed \textit{JointPGM}. The corresponding graphical representations of their overall computational paths are shown in Figure \ref{figure 2}. 

Normalization-based methods utilize adaptively learned mean $\bm{\mu}_t$ and variance $\bm{\sigma}^2_t$ to normalize the input observations $\bm{X}_{t-L:t}$ and encode them into their latent variable $\bm{Z}_{t-L:t}$, which is subsequently decoded into output predictions $\bm{X}_{t:t+H}$ through de-normalization, alleviating the temporal \textit{mean and covariance shift} between inputs and outputs. As Figure \ref{fig2_1} shows, this process assumes that the dependency between $\bm{X}_{t-L:t}$ and $\bm{X}_{t:t+H}$ (\textit{i.e.}, transitional distribution) remain fixed over time $t$.
As shown in Figure \ref{fig2_2}, time-variant models model the dependency between $\bm{X}_{t-L:t}$ and $\bm{X}_{t:t+H}$ at each time step $t$ (\textit{i.e.}, time-variant transitional distribution). However, this coarse-grained process mixes the transitional patterns occurring within each series and among different series, failing to reveal the intrinsic causes of the distribution shift. 

In contrast, \textit{JointPGM} segments $\bm{X}_{t-L:t}$ along the variable dimension, obtaining $N$ distinct series as input, and then encodes each series $\bm{x}_{t-L:t}^{(i)}$ separately into its latent variable $\hat{\bm{z}}_{t-L:t}^{(i)}$, as shown in Figure \ref{fig2_3}. The time factors $\bm{t}$ are additionally introduced to dynamically regulate the mapping processes both within each series ($\bm{t}, \bm{x}_{t-L:t}^{(i)} \rightarrow \hat{\bm{z}}_{t-L:t}^{(i)}$) and among different series ($\hat{\bm{z}}_{t-L:t}^{(i)} \rightarrow \bm{A}_t$). Thus, the intra-/inter-series correlations are captured, collectively forming the final latent variable $\bm{Z}_{t-L:t}$. In alignment with this process, $\bm{t} \rightarrow \bm{Z}_t$ denotes encoding time factors into the corresponding latent variable. Afterward, $\bm{Z}_{t-L:t} \rightarrow \bm{Z}_t$ means using a variational distribution $\mathcal{P}(\bm{Z}_t|\bm{Z}_{t-L:t})$ to approximate the distribution $\mathcal{P}(\bm{Z}_t|\bm{t})$. Herein, this relationship is designed to \textit{reversely infer} time factors in latent space. As the latent variable $\bm{Z}_{t-L:t}$ is exploited to generate $\bm{X}_{t:t+H}$, the time-variant transitional distribution is naturally decomposed into intra-/inter-series transitional distributions at a finer granularity.  

\subsection{Dual-Encoder Architecture} \label{sec3.3}
\textit{JointPGM} focuses on a probabilistic manner to account for the underlying causes of distribution shift in MTS forecasting. 
As Figure \ref{figure 3} shows, \textit{JointPGM} is organized with a dual-encoder architecture, which mainly involves four main components: 
1) \textit{Time factor encoder (TFE)} takes temporal order set $\{t-L+1,...,t+H\}$ as input to learn the dynamic time factors $\bm{M}_t^{(1)}$ and their latent variable $\bm{Z}_t$, which can reflect the clues of environmental changes; 
2) \textit{Independence-based series encoder (ISE)} captures series correlations by two distinct learners. While intra-series learner (left part of ISE in Figure \ref{figure 3}) focuses on capturing temporal dynamics within each series with temporal gate $\bm{G}^{(i)}$ adjusted by $\bm{M}_t^{(1)}$, inter-series learner (right part of ISE in Figure \ref{figure 3}) is to explicitly model the spatial dynamics with multi-hop propagation incorporating Gumbel-softmax sampling;
3) \textit{Dynamic Inference (DI)} uses latent variable $\bm{Z}_{t-L:t}$ to dynamically infer time factors and align with $\bm{Z}_t$;
4) \textit{Decoder} transforms $\bm{Z}_{t-L:t}$, formed by these two dynamics, into the final prediction and reconstruction.

\subsubsection{Time Factor Encoder (TFE)} 
Learning time factor representation that can accurately reflect irregular environmental changes is crucial for modeling distribution shifts. Transformer-based methods~\cite{DBLP:conf/aaai/ZhouZPZLXZ21,DBLP:conf/nips/WuXWL21,DBLP:conf/iclr/NieNSK23,DBLP:conf/iclr/ZhangY23} obtain learnable additive position encoding by heuristic sinusoidal mapping to distinguish the temporal order of tokens or patches. However, this design only monitors the temporal order of the lookback window, neglecting the association with its corresponding horizon window and thereby compromising predictive performance.  
In this regard, we propose to use temporal orders that span across both windows $\bm{t}=\{0,...,\frac{i+L}{L+H-1},...,1\}$ for $i=-L,-L+1,...,H-1$, \textit{i.e.}, a $[0,1]$-normalized temporal order set.
It is noteworthy that timestamp features (\textit{e.g.}, Minute-of-Hour, Day-of-Week, etc.) are also informative and can contribute to learning time factors. We opt for order features due to their more compact representations compared to timestamps. Additionally, embedding timestamp features with MLPs may have limitations in learning high-frequency patterns, commonly known as `spectral bias'~\cite{DBLP:conf/nips/TancikSMFRSRBN20,DBLP:conf/icml/WooLSKH23}. 

To obtain the high-quality representation of conditional information, we concatenate multiple Fourier basis functions with diverse scale parameters as suggested by~\cite{DBLP:conf/icml/WooLSKH23}, and then learn the deep features and align the dimensions using a feedforward neural network: 
\begin{equation} \label{eqn10}
    \bm{M}_t^{(0)}=\sin(2\pi \bm{B}_1 \bm{t})|\cos(2\pi \bm{B}_1 \bm{t})|...|\sin(2\pi \bm{B}_s \bm{t})|\cos(2\pi \bm{B}_s \bm{t}),
\end{equation}
\begin{equation} \label{eqn11}
    \bm{M}^{(1)}_t=\text{FeedForward}(\bm{M}^{(0)}_t),
\end{equation}
where elements in $\bm{B}_s \in \mathbb{R}^{\frac{b}{2s}}$ are sampled from $\mathcal{N}(0,\sigma^2_s)$ with $b$ denotes the Fourier feature size. $\bm{M}_t^{(0)} \in \mathbb{R}^{(L+H) \times b}$ and $\bm{M}_t^{(1)} \in \mathbb{R}^{L \times d}$ with $d$ denotes the latent dimension size. $\sigma_s \in \{0.01,0.1,1,5,10,20,$ $50,100\}$ denotes the scale hyperparameter and $s$ is its corresponding index starting from 1. 
$\cdot|\cdot$ represents the concatenation operation. ${\rm{FeedForward}}: \mathbb{R}^{(L+H) \times b} \rightarrow \mathbb{R}^{L \times d}$ is implemented by two linear layers with intermediate ReLU non-linearity. 
As shown in Figure \ref{figure 3}, taking the Fourier basis function $\cos(\cdot)$ as an example, its output has two main properties that could aid \textit{JointPGM} in distinguishing different temporal orders: similar temporal orders yield similar representations (\textit{e.g.}, the plot of $t,t+1$) and the larger the temporal order the earlier the values in representations oscillate between $-1$ and $+1$ (\textit{e.g.}, the plot of $t,t+H$).

Then, we model $\mathcal{P}(\bm{Z}_t|\bm{t})$ by stochastically sample $\bm{Z}_t$ from the Gaussian distribution using the reparameterization trick:
\begin{equation} \label{eqn15_1}
    \bm{\mu}_t=f^{\mu}_t(\bm{M}_t^{(1)}),
\end{equation}
\begin{equation} \label{eqn15_2}
    \bm{\sigma}_t^2=f^{\sigma^2}_t(\bm{M}_t^{(1)}),
\end{equation}
\begin{equation} \label{eqn15_3}
    \mathcal{P}(\bm{Z}_t|\bm{t})=\mathcal{N}(\bm{\mu}_t,\bm{\sigma}^2_t\bm{I}),
\end{equation}
where two multivariate functions $f^{\mu}_t(\cdot)$ and $f^{\sigma^2}_t(\cdot)$ map the input $\bm{M}_t^{(1)}$ to the mean and variance vectors of size $N \times d$ and $N \times d$. In practice, $f^{\mu}_t(\cdot)$ and $f^{\sigma^2}_t(\cdot)$ are instantiated as a single linear layer.

\subsubsection{Independence-based Series Encoder (ISE)} \label{sec3.B.2}
Series independence mechanism refers to the case of taking only one individual series as model input at each instance and mapping it into a latent space, rather than simultaneously incorporating all time series to mix information. 
This mechanism allows the model to only focus on learning information along the time axis and has shown effectiveness in working with linear models~\cite{DBLP:conf/aaai/ZengCZ023,DBLP:conf/nips/YiZFWWHALCN23} and Transformer-based models~\cite{DBLP:conf/iclr/NieNSK23} in time series forecasting tasks.
Therefore, we apply the series independence mechanism to sequentially explore two distinct types of correlations: intra- and inter-series correlations, thereby benefiting the modeling of time-variant transitional distribution within each series (\textit{i.e.}, $\mathcal{P}(\bm{x}_{t:t+H}^{(i)}|\bm{x}_{t-L:t}^{(i)},\bm{t})$) and among different series (\textit{i.e.}, $\mathcal{P}(\bm{x}_{t:t+H}^{(i)}|\bm{x}_{t-L:t}^{(\overline{i})},\bm{t})$) respectively. 
Note that such sequential style is beneficial and widely adopted by \cite{DBLP:conf/nips/YiZFWWHALCN23,DBLP:conf/kdd/EkambaramJNSK23}, with intra-series learner providing a solid representation foundation for inter-series learner.
ISE mainly consists of a intra-series learner and an inter-series learner, which are introduced as follows: 

\textbf{Intra-series Learner.}
The intra-series learner takes $\bm{X}_{t-L:t } \in \mathbb{R}^{N \times L}$ as input, which can be split into $N$ series. To illustrate the modeling process of intra-series transitional distribution, we draw inspiration from the structure of~\cite{DBLP:conf/aaai/ZengCZ023} and take the $i^{th}$ series $\bm{x}^{(i)}_{t-L:t} \in \mathbb{R}^{L}$ as an example. Concretely, $\bm{x}^{(i)}_{t-L:t}$ is fed into a linear layer according to our series-independent setting, then the linear layer will provide mapping results accordingly:
\begin{equation} \label{eqn8}
    \bm{h}^{(i)}_{t-L:t}=\text{Linear}(\bm{x}^{(i)}_{t-L:t}),
\end{equation}
where ${\rm{Linear}}: \mathbb{R}^L \rightarrow \mathbb{R}^d$ is implemented by a single linear layer, and $\bm{h}^{(i)}_{t-L:t} \in \mathbb{R}^{d}$. All series share the weights along the time dimension.

To render the modeled transitional distribution change over time, we design a temporal gate that is capable of distilling discriminative historical signals~\cite{10520822} sensitive to non-stationary environments based on dynamic time factors. Specifically,
we utilize a linear layer with a Sigmoid activation function to learn the temporal gate $\bm{G} \in \mathbb{R}^{N \times d}$. Subsequently, the $i^{th}$ gate $\bm{G}^{(i)}$ is applied to the representation $\bm{h}_{t-L:t}^{(i)}$ of the $i^{th}$ series: 
\begin{equation} \label{eqn9_1}
    \bm{G}=\text{Sigmoid}(\text{Linear}(\bm{M}_t^{(1)})),
\end{equation}
\begin{equation} \label{eqn9_2}
    \hat{\bm{h}}_{t-L:t}^{(i)}=\bm{G}^{(i)} \odot \bm{h}_{t-L:t}^{(i)},
\end{equation}
where $\odot$ is element-wise multiplication, and all sub-gates $\bm{G}^{(i)},i \in \{1,...,N\}$ share the weights along time dimension.

To capture the transitional distribution for each series, we explicitly model the distribution $\mathcal{P}(\hat{\bm{z}}_{t-L:t}^{(i)}|\bm{x}_{t-L:t}^{(i)},\bm{t})$ by stochastically sampling each latent variable $\hat{\bm{z}}_{t-L:t}^{(i)}$ from a Gaussian distribution: 
\begin{equation} \label{eqn14_1}
     \bm{\mu}_{\hat{z}}=f^{\mu}_{\hat{z}}(\hat{\bm{h}}^{(i)}_{t-L:t}),
\end{equation}
\begin{equation} \label{eqn14_2}
     \bm{\sigma}^2_{\hat{z}}=f^{\sigma^2}_{\hat{z}}(\hat{\bm{h}}^{(i)}_{t-L:t}),
\end{equation}
\begin{equation} \label{eqn14_3}
     \mathcal{P}(\hat{\bm{z}}_{t-L:t}^{(i)}|\bm{x}_{t-L:t}^{(i)},\bm{t})=\mathcal{N}(\bm{\mu}_{\hat{z}},\bm{\sigma}^2_{\hat{z}}\bm{I}),
\end{equation}
where $\hat{\bm{z}}_{t-L:t}^{(i)} \in \mathbb{R}^{d}$. $f^{\mu}_{\hat{z}}(\cdot)$ and $f^{\sigma^2}_{\hat{z}}(\cdot)$ are instantiated as a single linear layer and share weights between the latent states of $N$ series. Finally, we ensemble $\hat{\bm{h}}^{(i)}_{t-L:t}$ and $\hat{\bm{z}}^{(i)}_{t-L:t}$ of $N$ time series into a whole respectively, yielding respective outputs $\hat{\bm{H}}_{t-L:t} \in \mathbb{R}^{N \times d}$ and $\hat{\bm{Z}}_{t-L:t} \in \mathbb{R}^{N \times d}$. Considering the previously implemented series-independent processes, where each series $\bm{x}_{t-L:t}^{(i)}$ is independent of each other series belonging to  $\bm{x}_{t-L:t}^{(\overline{i})}$, we can compose the posterior distribution $\mathcal{P}(\hat{\bm{Z}}_{t-L:t}|\bm{X}_{t-L:t},\bm{t})$ from multiple sub-distributions $\mathcal{P}(\hat{\bm{z}}_{t-L:t}^{(i)}|\bm{x}_{t-L:t}^{(i)},\bm{t}),\ i \in \{1,...,N\}$.

\textbf{Inter-series Learner.} 
Most methods~\cite{DBLP:conf/kdd/WuPL0CZ20,DBLP:conf/nips/0001YL0020,DBLP:conf/aaai/YangLWZPW23} randomly initialize node embeddings for all nodes and infer the dependencies between each pair of nodes by multiplication operations. The adjacency matrix derived in this way is essentially input-unconditioned, making it challenging to effectively handle abrupt changes in non-stationary time series. Hence, we propose to calculate the relationships between nodes by the self-attention mechanism~\cite{DBLP:conf/nips/CaoWDZZHTXBTZ20}:
\begin{equation} \label{eqn1_1}
    \bm{Q}_t=\hat{\bm{Z}}_{t-L:t}\bm{W}^{\bm{Q}}_t,
\end{equation}
\begin{equation} \label{eqn1_2}
    \bm{K}_t=\hat{\bm{Z}}_{t-L:t}\bm{W}^{\bm{K}}_t,
\end{equation}
\begin{equation} \label{eqn1_3}
    \bm{W}_t=\text{Softmax}(\frac{\bm{Q}_t\bm{K}_t^{\rm T}}{\sqrt{d}}),
\end{equation}
where $\bm{Q}_t$ and $\bm{K}_t$ indicate the representation for query and key at time step $t$, which can be calculated by linear projections with learnable parameters $\bm{W}_t^{\bm{Q}}$ and $\bm{W}_t^{\bm{K}}$ respectively. Here, $\bm{W}_t$ is the continuous version of the adjacency matrix (\textit{a.k.a.}, probability matrix) then $w_{ij,t} \in \bm{W}_t$ denotes the probability to preserve the edge of series $i$ to $j$ at time step $t$. However, such soft weights are incapable of decisively choosing between retaining or discarding edges, thereby hindering the explicit modeling of how each series is influenced by its relevant series during the distribution shift. 
Therefore, inspired by~\cite{DBLP:conf/iclr/Shang0B21,DBLP:conf/ijcai/YuLYLHWL22}, we apply the Gumbel reparameterization trick:
\begin{equation} \label{eqn2}
  \begin{aligned}
    a_{ij,t}=\text{Sigmoid}&((\log(w_{ij,t}/(1-w_{ij,t})))+(g^1_{ij,t}-g^2_{ij,t}))/\tau), \\ 
    &s.t. \ \  g^1_{ij,t},g^2_{ij,t} \sim \text{Gumbel}(0,1),
  \end{aligned}
\end{equation}
where $\tau \in (0,\infty)$ is a temperature parameter. When $\tau \rightarrow 0, a_{ij,t}=1 \in \bm{A}_t$ with probability $w_{ij,t}$ and 0 with remaining probability. 

Afterward, we utilize multi-hop propagation, which is a simplified version of mix-hop propagation proposed by~\cite{DBLP:conf/kdd/WuPL0CZ20,DBLP:conf/aaai/YangLWZPW23}, to aggregate information from immediate neighbors. Given the input $\hat{\bm{H}}_{t-L:t}$ and adjacency matrix $\bm{A}_t$, the process of $K$-layer propagation can be formulated as follows:
\begin{equation} \label{eqn13_1}
    \tilde{\bm{H}}_{t-L:t}^{(k)}=\bm{A}_t\tilde{\bm{H}}_{t-L:t}^{k-1},
\end{equation}
\begin{equation} \label{eqn13_2}
    \tilde{\bm{H}}_{t-L:t}=\sum_{k=1}^{K}\text{Linear}^{(k)}(\tilde{\bm{H}}_{t-L:t}^{(k)}),
\end{equation}
where $K$ is the depth of propagation, $\tilde{\bm{H}}_{t-L:t} \in \mathbb{R}^{N \times d}$ denotes the output representation of the current layer, $\tilde{\bm{H}}_{t-L:t}^{(0)}=\hat{\bm{H}}_{t-L:t}$.
Such simplification provides an important insight: under the wild non-stationarity, mixing the original representation $\hat{\bm{H}}_{t-L:t}$ in each hop can easily introduce noise into the inter-series correlation learning, as well as into the subsequent inter-series transitional distribution modeling. 

Here we explicate the distribution modeling from a holistic perspective. We denote the latent state of $\bm{X}_{t-L:t}$ that are inferred from $\tilde{\bm{H}}_{t-L:t}$ by $\tilde{\bm{Z}}_{t-L:t}$, which is distinguished from $\hat{\bm{Z}}_{t-L:t}$ achieved by the temporal gate. The distribution $\mathcal{P}(\tilde{\bm{Z}}_{t-L:t}|\bm{X}_{t-L:t},\bm{t})$ is stochastically sampled $\tilde{\bm{Z}}_{t-L:t}$ from the Gaussian distribution: 
\begin{equation} \label{eqn16_1}
     \bm{\mu}_{\tilde{Z}}=f^{\mu}_{\tilde{Z}}(\tilde{\bm{H}}_{t-L:t}),
\end{equation}
\begin{equation} \label{eqn16_2}
     \bm{\sigma}^2_{\tilde{Z}}=f^{\sigma^2}_{\tilde{Z}}(\tilde{\bm{H}}_{t-L:t}),
\end{equation}
\begin{equation} \label{eqn16_3}
     \mathcal{P}(\tilde{\bm{Z}}_{t-L:t}|\bm{X}_{t-L:t},\bm{t})=\mathcal{N}(\bm{\mu}_{\tilde{Z}},\bm{\sigma}^2_{\tilde{Z}}\bm{I}),
\end{equation}
where $\tilde{\bm{Z}}_{t-L:t} \in \mathbb{R}^{N \times d}$, $f^{\mu}_{\tilde{Z}}(\cdot)$ and $f^{\sigma^2}_{\tilde{Z}}(\cdot)$ are also instantiated as a single linear layer. For each series $\tilde{\bm{z}}_{t-L:t}^{(i)} \in \tilde{\bm{Z}}_{t-L:t}$, the posterior approximation can be successfully represented by a product of two sub-distributions: $\mathcal{P}(\tilde{\bm{z}}_{t-L:t}^{(i)}|\bm{x}_{t-L:t}^{(\overline{i})},\bm{t})=\mathcal{P}(\tilde{\bm{z}}^{(i)}_{t-L:t}|\hat{\bm{z}}^{(\overline{i})}_{t-L:t})\mathcal{P}(\hat{\bm{z}}^{(\overline{i})}_{t-L:t}|\bm{x}^{(\overline{i})}_{t-L:t},\bm{t})$. Accordingly, the process of Gumbel-softmax sampling explicitly selects correlated series $\bm{x}^{(\overline{i})}_{t-L:t}$ for each series $\bm{x}_{t-L:t}^{(i)}, i \in \{1,...,N\}$, which is part of the former, while the latter involves the independent modeling of each correlated series.



Then, the latent variable output from intra-series learner, $\hat{\bm{Z}}_{t-L:t}$, joints the latent variable output from inter-series learner, $\tilde{\bm{Z}}_{t-L:t}$, to form $\bm{Z}_{t-L:t} \sim \mathcal{P}(\bm{Z}_{t-L:t}|\bm{X}_{t-L:t},\bm{t})$, with their proportions regulated by trade-off parameter $\alpha$:
\begin{equation} \label{eqn17}
     \bm{Z}_{t-L:t}=\alpha\hat{\bm{Z}}_{t-L:t}+(1-\alpha)\tilde{\bm{Z}}_{t-L:t},
\end{equation}

\subsubsection{Dynamic Inference (DI)}
As illustrated in Section \ref{sec3.2}, we aim to use variational distribution $\mathcal{P}(\hat{\bm{Z}}_t|\bm{Z}_{t-L:t})$ to estimate the distribution $\mathcal{P}(\bm{Z}_t|\bm{t})$ achieved by time factor encoder. Accordingly, we derive the latent variable $\hat{\bm{Z}}_t$ through a linear layer. After that, we use two linear functions $f^{\mu}_{\hat{t}}(\cdot)$ and $f^{\sigma^2}_{\hat{t}}(\cdot)$ to map the latent state $\hat{\bm{Z}}_t$ to the mean and variance vectors, as formulated follows:
\begin{equation} \label{eqn18_1}
     \hat{\bm{Z}}_t={\rm{Linear}}(\bm{Z}_{t-L:t}),
\end{equation}
\begin{equation} \label{eqn18_2}
     \bm{\mu}_{\hat{t}}=f^{\mu}_{\hat{t}}(\hat{\bm{Z}}_t),
\end{equation}
\begin{equation} \label{eqn18_3}
     \bm{\sigma}^2_{\hat{t}}=f^{\sigma^2}_{\hat{t}}(\hat{\bm{Z}}_t),
\end{equation}

\subsubsection{Decoder}
We utilize the learned latent variable $\bm{Z}_{t-L:t}$ 
to perform reconstruction and prediction with one forward step which can avoid error accumulation, as formulated below:
\begin{equation} \label{eqn19_1}
     \hat{\bm{X}}_{t-L:t}={\rm{FeedForward}}_{rec}(\bm{Z}_{t-L:t}),
\end{equation}
\begin{equation} \label{eqn19_2}
     \hat{\bm{X}}_{t:t+H}={\rm{FeedForward}}_{pre}(\bm{Z}_{t-L:t}),
\end{equation}
where ${\rm{FeedForward}}_{rec}: \mathbb{R}^d \rightarrow \mathbb{R}^L$ and ${\rm{FeedForward}}_{pre}: \mathbb{R}^d \rightarrow \mathbb{R}^H$ are both implemented using two linear layers with intermediate LeakyReLU non-linearity.

\subsection{Objective Decomposition} \label{sec3.4}
For simplicity, we abbreviate $\bm{X}_{t-L:t}$ as $\bm{X}_L$, $\bm{Z}_{t-L:t}$ as $\bm{Z}_L$, $\bm{X}_{t:t+H}$ as $\bm{X}_H$, and omit intermediate variables $\hat{\bm{z}}_{t-L:t}^{(i)}$ and $\bm{A}_t$ when there is no confusion.
To tackle the distribution shift in MTS forecasting, our objective is to explicitly model the time-variant transitional distribution between output predictions and input observations. 
It requires the learned latent variables of time series to be informative and discriminative, while also exhibiting a high sensitivity to dynamic time factors that can reflect non-stationary environments.
Therefore, based on our tailored PGM, we conduct the following variational inference using the Kullback-Leibler ($\mathbb{KL}$) divergence:
\begin{equation} \label{eqn3}
    \mathcal{L}=\mathbb{KL}[\mathcal{P}_\psi(\bm{X}_H,\bm{Z}_L,\bm{Z}_t|\bm{X}_L)||\mathcal{P}_\phi(\bm{X}_H,\bm{Z}_L,\bm{Z}_t|\bm{X}_L,\bm{t})],
\end{equation}
where $\mathcal{P}_\psi(\cdot|\cdot)$ and $\mathcal{P}_\phi(\cdot|\cdot)$ are two transitional distributions, with $\psi$ and $\phi$ denoting the parameterized functions. The divergence Eq. (\ref{eqn3}) is minimized concerning all parameters. 

\begin{proposition}
    \label{pro1}
    Regarding the $\mathbb{KL}$ divergence Eq. (\ref{eqn3}), we show that the divergence can be decomposed as: 
        \begin{align}
        \label{eqn4} \nonumber
        \mathcal{L}&=\mathbb{KL}[\mathcal{P}_\psi(\bm{X}_H,\bm{Z}_L,\bm{Z}_t|\bm{X}_L)||\mathcal{P}_\phi(\bm{X}_H,\bm{Z}_L,\bm{Z}_t|\bm{X}_L,\bm{t})] \\  \nonumber
        &=\underbrace{\mathbb{KL}[\mathcal{P}_\psi(\bm{Z}_L|\bm{X}_L)||\mathcal{P}_\phi(\bm{Z}_L|\bm{X}_L,\bm{t})]}_{(a)} \\  
        &+\underbrace{\mathbb{E}_{\bm{Z}_L \sim \mathcal{P}_\psi(\bm{Z}_L|\bm{X}_L)}\mathbb{KL}[\mathcal{P}_\theta(\bm{Z}_t|\bm{Z}_L)||\mathcal{P}_\phi(\bm{Z}_t|\bm{t})]}_{(b)}  \\  \nonumber
        &+\underbrace{\mathbb{E}_{(\bm{Z}_L,\bm{Z}_t) \sim \mathcal{P}_\psi(\bm{Z}_L,\bm{Z}_t|\bm{X}_L)}\mathbb{KL}[\mathcal{P}_\psi(\bm{X}_H|\bm{Z}_L)||\mathcal{P}_\phi(\bm{X}_H|\bm{X}_L,\bm{t})]}_{(c)},  \nonumber
        \end{align}
\end{proposition}

We prove the Proposition \ref{pro1} in Appendix \ref{secB.1}. Then, we detailedly analyze the above three terms in Eq. (\ref{eqn4}) combining the designed PGM and our purpose, and provide the loss function for each item and overall loss function.

For term ($a$), \textit{i.e.}, $\mathbb{KL}[\mathcal{P}_{\psi}(\bm{Z}_L|\bm{X}_L)||\mathcal{P}_{\phi}(\bm{Z}_L|\bm{X}_L,\bm{t})]$, it aims to keep the inference using the variational distribution and the inference using the posterior is \textit{close}, which also guarantees the reliable and high-quality sampling. We employ the variational evidence lower bound (ELBO) to constrain the term ($a$). Mathematically, we have: 
\begin{equation} \label{eqn21}
    \begin{split}
    \mathcal{L}_a&=-ELBO \\
    &=-\mathbb{E}_{\bm{Z}_L}[\log\mathcal{P}(\bm{X}_L|\bm{Z}_L)]+\mathbb{KL}[\mathcal{P}_{\psi}(\bm{Z}_L|\bm{X}_L)||\mathcal{P}(\bm{Z}_L)] \\ 
    &=l(\hat{\bm{X}}_L-\bm{X}_L) + (-\log\bm{\sigma}_Z+\frac{1}{2}\bm{\sigma}^2_Z+\frac{1}{2}\bm{\mu}_Z^2-\frac{1}{2}),
    \end{split}
\end{equation}
where $l$ denotes a distance metric for which we use the MSE loss, 
$\bm{\mu}_{Z}$ and $\bm{\sigma}^2_Z$ are the mean and variance vectors of $\bm{Z}_L$. 

For term (b), \textit{i.e.}, $\mathbb{E}_{\bm{Z}_L \sim \mathcal{P}_\psi(\bm{Z}_L|\bm{X}_L)}\mathbb{KL}[\mathcal{P}_\theta(\bm{Z}_t|\bm{Z}_L)||\mathcal{P}_\phi(\bm{Z}_t|\bm{t})]$, to make time factors more sensitive to non-stationary environments, we use a variational distribution to approximate posterior distribution $\mathcal{P}_\phi(\bm{Z}_t|\bm{t})$. Mathematically, we have:
\begin{equation} \label{eqn20}
    \mathcal{L}_b=-\log\frac{\bm{\sigma}_{\hat{t}}}{\bm{\sigma}_t}+\frac{1}{2}\frac{\bm{\sigma}^2_{\hat{t}}}{\bm{\sigma}^2_t}+\frac{1}{2}\frac{(\bm{\mu}_{\hat{t}}-\bm{\mu}_t)^2}{\bm{\sigma}^2_t}-\frac{1}{2},      
\end{equation}
where $\bm{\mu}_t$ and $\bm{\sigma}^2_t$ are the mean and variance of $\bm{Z}_t$. $\bm{\mu}_{\hat{t}}$ and $\bm{\sigma}^2_{\hat{t}}$ are the mean and variance of $\hat{\bm{Z}}_t$.

For term (c), \textit{i.e.}, $\mathbb{E}_{(\bm{Z}_L,\bm{Z}_t) \sim \mathcal{P}_\psi(\bm{Z}_L,\bm{Z}_t|\bm{X}_L)}\mathbb{KL}[\mathcal{P}_\psi(\bm{X}_H|\bm{Z}_L)||$ $\mathcal{P}_\phi(\bm{X}_H|\bm{X}_L,\bm{t})]$, we can generate $\bm{X}_H$ with $\bm{X}_L$ and $\bm{t}$ or with the latent variable $\bm{Z}_L$. We minimize the distance about the generation of $\bm{X}_H$ in two ways. The minimization can ensure that the generation from the raw series/time and latent variables is \textit{consistent}.
To ensure the minimization of the term (c) in Eq. (\ref{eqn4}), we consider integrating the use of forecasting and reconstruction losses, where reconstruction loss is omitted as it has already been constrained in term (a):
\begin{equation} \label{eqn7}
    \mathcal{L}_c=l(\hat{\bm{X}}_H-\bm{X}_H),       
\end{equation}

Finally, the overall loss function is the sum of the above three losses:
\begin{equation} \label{eqn23}
    \mathcal{L}=\mathcal{L}_a+\mathcal{L}_b+\mathcal{L}_c.     
\end{equation}
The derivation of Eq. (\ref{eqn21}) is presented in Appendix \ref{secB.2}. Further discussion on \textit{JointPGM} is provided in Appendix \ref{sece}.

\begin{table}[!t]
 \centering
 \caption{Summary of Datasets. A larger ADF test statistic means a higher level of non-stationarity, \textit{i.e.}, a more severe distribution shift.}
 \label{table C.1}
 \newcommand{\tabincell}[2]{\begin{tabular}{@{}#1@{}}#2\end{tabular}}
 \begin{tabular}{ccccc}
  \toprule
  Datasets & Variable & Interval & Time Step & ADF Test Statistic \\
  \midrule 
  Exchange & 8 & 1 day & 7588 & -1.902 \\
  \midrule 
  ETTm2 & 7 & 15 minutes & 69680 & -5.664 \\
  \midrule 
  ETTh1 & 7 & 1 hour & 17420 & -5.909 \\
  \midrule 
  Electricity & 321 & 1 hour & 26304 & -8.445 \\  
  \midrule 
  METR-LA & 207 & 5 minutes & 34272 & -15.021 \\
  \midrule
  ILI & 7 & 1 week & 966 & -5.334   \\
  \bottomrule 
 \end{tabular}
\end{table}


\begin{table*}[!t]
\setlength{\tabcolsep}{3.3pt}
 \centering
 \caption{Forecasting result comparison with different horizon lengths. The lookback length is set to 24 for ILI and 96 for the others. \textbf{Bold} indicates the best result, while \underline{underlining} indicates the second-best result.}
 \label{table 1}
 \newcommand{\tabincell}[2]{\begin{tabular}{@{}#1@{}}#2\end{tabular}}
 \scalebox{0.97}{
 \begin{tabular}{c|c|cccccccccccccccccccc}
  \toprule
  \multicolumn{2}{c}{Models} & \multicolumn{2}{c}{\textbf{JointPGM}} & \multicolumn{2}{c}{Koopa} & \multicolumn{2}{c}{Stationary} & \multicolumn{2}{c}{DLinear} & \multicolumn{2}{c}{PatchTST} & \multicolumn{2}{c}{FEDformer} & \multicolumn{2}{c}{Autoformer} & \multicolumn{2}{c}{iTransformer} & \multicolumn{2}{c}{Crossformer} & \multicolumn{2}{c}{WaveForM} \\
  \cmidrule(r){3-4} \cmidrule(r){5-6} \cmidrule(r){7-8} \cmidrule(r){9-10} \cmidrule(r){11-12} \cmidrule(r){13-14} \cmidrule(r){15-16} \cmidrule(r){17-18} \cmidrule(r){19-20} \cmidrule(r){21-22}
  \multicolumn{2}{c}{Metrics} & MAE & MSE & MAE & MSE & MAE & MSE & MAE & MSE & MAE & MSE & MAE & MSE & MAE & MSE & MAE & MSE & MAE & MSE & MAE & MSE  \\ 
  \midrule 
  \multirow{4}{*}{\rotatebox{90}{Exchange}} & 48 & \textbf{0.143} & \textbf{0.042} & 0.154 & 0.049 & 0.181 & 0.064 & 0.154 & 0.046 & \underline{0.149} & 0.047 & 0.234 & 0.103 & 0.238 & 0.105 & \underline{0.149} & \underline{0.045} & 0.272 & 0.137 & 0.454 & 0.405 \\
  & 96 & \textbf{0.199} & \textbf{0.076} & 0.214 & 0.093 & 0.281 & 0.151 & 0.215 & 0.087 & 0.208 & 0.091 & 0.286 & 0.156 & 0.275 & 0.145 & \underline{0.206} & \underline{0.086} & 0.397 & 0.269 & 0.597 & 0.688 \\
  & 192 & \textbf{0.293} & \textbf{0.155} & 0.304 & 0.180 & 0.347 & 0.222 & 0.305 & \underline{0.167} & \underline{0.298} & 0.168 & 0.380 & 0.271 & 0.383 & 0.273 & 0.303 & 0.180 & 0.489 & 0.395 & 0.775 & 1.079 \\
  & 336 & \textbf{0.398} & \textbf{0.283} & 0.440 & 0.365 & 0.460 & 0.385 & \underline{0.416} & \underline{0.303} & 0.419 & 0.341 & 0.490 & 0.440 & 0.495 & 0.444 & 0.421 & 0.339 & 0.727 & 0.859 & 0.983 & 1.631 \\
  \midrule 
  \multirow{4}{*}{\rotatebox{90}{ETTm2}} & 48 & 0.215 & \textbf{0.101} & \underline{0.213} & \textbf{0.101} & 0.217 & \underline{0.102} & 0.218 & \underline{0.102} & 0.217 & 0.104 & 0.220 & \underline{0.102} & 0.263 & 0.141 & \textbf{0.212} & \underline{0.102} & 0.244 & 0.118 & 0.222 & 0.155 \\
  & 96 & \textbf{0.235} & \textbf{0.121} & 0.238 & 0.123 & 0.249 & 0.140 & \underline{0.237} & 0.123 & 0.240 & \textbf{0.121} & 0.239 & \textbf{0.121} & 0.258 & 0.142 & 0.238 & 0.125 & 0.245 & \underline{0.122} & 0.244 & 0.124 \\
  & 192 & \textbf{0.261} & \textbf{0.149} & \underline{0.263} & \underline{0.150} & 0.279 & 0.170 & 0.265 & \underline{0.150} & 0.296 & 0.176 & \underline{0.263} & 0.157 & 0.271 & 0.157 & 0.268 & 0.155 & 0.280 & 0.155 & 0.296 & 0.231 \\
  & 336 & \textbf{0.289} & \textbf{0.179} & 0.292 & 0.185 & 0.312 & 0.215 & \underline{0.291} & \underline{0.180} & 0.325 & 0.206 & 0.293 & 0.184 & 0.300 & 0.190 & 0.299 & 0.192 & 0.322 & 0.207 & 0.336 & 0.284 \\
  \midrule 
  \multirow{4}{*}{\rotatebox{90}{ETTh1}} & 48 & \textbf{0.410} & \textbf{0.386} & 0.421 & 0.389 & 0.491 & 0.485 & \underline{0.414} & 0.391 & 0.419 & 0.394 & 0.482 & 0.446 & 0.526 & 0.542 & 0.416 & \underline{0.387} & 0.446 & 0.409 & 0.456 & 0.446 \\
  & 96 & \textbf{0.444} & \textbf{0.436} & 0.458 & 0.444 & 0.535 & 0.552 & \underline{0.451} & 0.450 & 0.455 & 0.438 & 0.514 & 0.493 & 0.583 & 0.653 & 0.456 & 0.441 & 0.453 & \underline{0.437} & 0.505 & 0.523 \\
  & 192 & \textbf{0.484} & \textbf{0.491} & 0.493 & 0.496 & 0.578 & 0.626 & \underline{0.489} & 0.504 & 0.495 & 0.496 & 0.554 & 0.556 & 0.596 & 0.654 & 0.499 & 0.501 & 0.508 & \underline{0.494} & 0.595 & 0.697 \\
  & 336 & \textbf{0.519} & \textbf{0.534} & 0.528 & \underline{0.553} & 0.627 & 0.706 & \underline{0.521} & 0.554 & 0.528 & 0.560 & 0.575 & 0.595 & 0.607 & 0.668 & 0.536 & 0.562 & 0.575 & 0.600 & 0.612 & 0.726 \\
  \midrule 
  \multirow{4}{*}{\rotatebox{90}{Electricity}} & 48 & 0.272 & 0.191 & \textbf{0.237} & \textbf{0.135} & 0.282 & 0.173 & 0.274 & 0.194 & \underline{0.248} & 0.158 & 0.313 & 0.199 & 0.318 & 0.205 & 0.250 & \underline{0.151} & 0.268 & 0.162 & 0.302 & 0.213 \\
  & 96 & \textbf{0.256} & \textbf{0.161} & 0.264 & \underline{0.164} & 0.303 & 0.201 & 0.277 & 0.195 & \underline{0.257} & 0.168 & 0.320 & 0.206 & 0.338 & 0.232 & 0.263 & 0.165 & 0.286 & 0.185 & 0.322 & 0.230 \\
  & 192 & \textbf{0.279} & \textbf{0.187} & 0.281 & 0.189 & 0.312 & 0.207 & \underline{0.280} & 0.193 & \underline{0.280} & \underline{0.188} & 0.328 & 0.214 & 0.356 & 0.249 & \textbf{0.279} & 0.189 & 0.298 & 0.203 & 0.351 & 0.255 \\
  & 336 & \underline{0.295} & \textbf{0.202} & 0.296 & 0.205 & 0.332 & 0.232 & 0.296 & 0.207 & \textbf{0.287} & 0.208 & 0.357 & 0.256 & 0.382 & 0.284 & 0.296 & \underline{0.203} & 0.324 & 0.234 & 0.381 & 0.288 \\
  \midrule 
  \multirow{4}{*}{\rotatebox{90}{METR-LA}} & 48 & 0.591 & \textbf{0.874} & 0.617 & 1.114 & 0.617 & 1.226 & \underline{0.589} & 0.884 & 0.626 & 1.118 & 0.696 & 1.147 & 0.813 & 1.581 & 0.609 & 1.114 & 0.590 & 0.907 & \textbf{0.576} & \underline{0.876} \\
  & 96 & \textbf{0.622} & \textbf{1.041} & 0.720 & 1.435 & 0.710 & 1.504 & 0.694 & 1.101 & 0.771 & 1.484 & 0.781 & 1.450 & 0.904 & 1.787 & 0.751 & 1.458 & \underline{0.680} & \underline{1.061} & 0.684 & 1.114 \\
  & 192 & \textbf{0.756} & \textbf{1.246} & 0.833 & 1.700 & 0.802 & 1.828 & \textbf{0.756} & \underline{1.248} & 0.852 & 1.702 & 0.864 & 1.746 & 0.965 & 1.993 & 0.842 & 1.739 & 0.768 & 1.307 & \underline{0.758} & 1.259 \\
  & 336 & \textbf{0.771} & \textbf{1.297} & 0.849 & 1.778 & 0.807 & 1.857 & 0.777 & \underline{1.301} & 0.857 & 1.762 & 0.954 & 1.999 & 0.954 & 2.019 & 0.863 & 1.829 & 0.796 & 1.364 & \underline{0.775} & 1.324 \\
  \midrule 
  \multirow{4}{*}{\rotatebox{90}{ILI}} & 24 & 1.354 & \underline{3.818} & \textbf{1.285} & \textbf{3.697} & 1.368 & 4.679 & 1.408 & 4.127 & \underline{1.333} & 4.000 & 1.714 & 5.447 & 1.716 & 5.449 & 1.437 & 4.525 & 1.424 & 4.458 & 1.977 & 7.189 \\
  & 36 & 1.390 & \textbf{3.893} & 1.364 & \underline{3.967} & \textbf{1.286} & 4.194 & \underline{1.359} & 3.983 & 1.414 & 4.432 & 1.584 & 4.853 & 1.606 & 4.865 & 1.428 & 4.513 & 1.698 & 5.538 & 2.356 & 9.444 \\
  & 48 & 1.340 & \textbf{3.640} & \underline{1.291} & 3.657 & \textbf{1.241} & 3.903 & 1.299 & \underline{3.650} & 1.378 & 4.184 & 1.461 & 4.310 & 1.432 & 4.261 & 1.420 & 4.447 & 1.684 & 5.702 & 2.228 & 8.886 \\
  & 60 & 1.372 & \textbf{3.830} & 1.394 & 4.312 & \textbf{1.276} & 3.966 & \underline{1.366} & \underline{3.897} & 1.378 & 4.229 & 1.504 & 4.519 & 1.491 & 4.414 & 1.451 & 4.554 & 1.603 & 5.302 & 2.294 & 9.161 \\
  \midrule 
  \multicolumn{2}{c}{1$^{st}$ Count} & \multicolumn{2}{c}{\textbf{38}} & \multicolumn{2}{c}{\underline{5}} & \multicolumn{2}{c}{3} & \multicolumn{2}{c}{1} & \multicolumn{2}{c}{2} & \multicolumn{2}{c}{1} & \multicolumn{2}{c}{0} & \multicolumn{2}{c}{2} & \multicolumn{2}{c}{0} & \multicolumn{2}{c}{1} \\
  \bottomrule 
 \end{tabular}
 }
\end{table*}

\begin{table*}[!t]
\setlength{\tabcolsep}{4pt}
 \centering
 \caption{Forecasting result comparison to SAN~\cite{DBLP:conf/nips/LiuCLHLXC23}, Dish-TS~\cite{DBLP:conf/aaai/0010WWWZF23} and RevIN~\cite{DBLP:conf/iclr/KimKTPCC22} with different backbones including Autoformer and FEDformer. The lookback/horizon length is set to $L/H=24/48$ for ILI and $L/H=96/96$ for the others. \textbf{Bold} indicates the best result, while \underline{underlining} indicates the second-best result. The comparison results for the return-to-original-value setting are shown in Appendix \ref{secd.1}, Table \ref{table C.3}.}
 \label{table c.2}
 \newcommand{\tabincell}[2]{\begin{tabular}{@{}#1@{}}#2\end{tabular}}
 \begin{tabular}{c|cccccccccccccccccc}
  \toprule
  Models & \multicolumn{2}{c}{\textbf{JointPGM}} & \multicolumn{2}{c}{Autoformer} & \multicolumn{2}{c}{+SAN} & \multicolumn{2}{c}{+Dish-TS} & \multicolumn{2}{c}{+RevIN} & \multicolumn{2}{c}{FEDformer} & \multicolumn{2}{c}{+SAN} & \multicolumn{2}{c}{+Dish-TS} & \multicolumn{2}{c}{+RevIN} \\
  \cmidrule(r){2-3} \cmidrule(r){4-5} \cmidrule(r){6-7} \cmidrule(r){8-9} \cmidrule(r){10-11} \cmidrule(r){12-13} \cmidrule(r){14-15} \cmidrule(r){16-17} \cmidrule(r){18-19} 
  Metrics & MAE & MSE & MAE & MSE & MAE & MSE & MAE & MSE & MAE & MSE & MAE & MSE & MAE & MSE & MAE & MSE & MAE & MSE \\ 
  \midrule 
  Exchange & \textbf{0.199} & \textbf{0.076} & 0.275 & 0.145 & 0.209 & 0.084 & 0.287 & 0.162 & 0.304 & 0.175 & 0.286 & 0.156 & \underline{0.206} & \underline{0.080} & 0.340 & 0.205 & 0.292 & 0.153 \\
  ETTm2 & \textbf{0.235} & \textbf{0.121} & 0.258 & 0.142 & 0.238 & \underline{0.124} & 0.259 & 0.136 & 0.257 & 0.140 & 0.239 & \textbf{0.121} & \underline{0.236} & \textbf{0.121} & 0.289 & 0.161 & 0.261 & 0.162 \\
  ETTh1 & \textbf{0.444} & \textbf{0.436} & 0.583 & 0.653 & 0.540 & 0.592 & 0.553 & 0.593 & 0.583 & 0.667 & 0.514 & 0.493 & \underline{0.473} & \underline{0.445} & 0.508 & 0.514 & 0.595 & 0.678 \\
  Electricity & \textbf{0.256} & \textbf{0.161} & 0.338 & 0.232 & 0.282 & 0.173 & 0.537 & 0.520 & 0.581 & 0.566 & 0.320 & 0.206 & \underline{0.272} & \underline{0.165} & 0.476 & 0.425 & 0.356 & 0.264  \\
  METR-LA & \textbf{0.622} & \textbf{1.041} & 0.904 & 1.787 & 0.659 & 1.132 & 0.748 & 1.314 & 0.745 & 1.461 & 0.781 & 1.450 & \underline{0.628} & \underline{1.044} & 0.764 & 1.290 & 0.756 & 1.460 \\
  ILI & 1.340 & \textbf{3.640} & 1.432 & 4.261 & 1.367 & \underline{3.825} & 1.395 & 4.181 & \underline{1.330} & 3.906 & 1.461 & 4.310 & 1.354 & 3.880 & 1.465 & 4.593 & \textbf{1.322} & 4.304 \\
  \midrule 
  1$^{st}$ Count & \multicolumn{2}{c}{\textbf{11}} & \multicolumn{2}{c}{0} & \multicolumn{2}{c}{0} & \multicolumn{2}{c}{0} & \multicolumn{2}{c}{0} & \multicolumn{2}{c}{1} & \multicolumn{2}{c}{1} & \multicolumn{2}{c}{0} & \multicolumn{2}{c}{1} \\
  \bottomrule 
 \end{tabular}
\end{table*}

\section{Experiments} \label{sec4}

\subsection{Experimental Setup}

\subsubsection{Datasets}
We conduct extensive experiments on various datasets
to evaluate the performance and efficiency of \textit{JointPGM}. We include six well-acknowledged benchmarks used in previous non-stationary time series forecasting works~\cite{DBLP:conf/iclr/KimKTPCC22,DBLP:conf/aaai/0010WWWZF23,DBLP:conf/nips/LiuWWL22,10509830,DBLP:conf/nips/LiuLWL23,DBLP:conf/nips/LiuCLHLXC23}: Exchange\footnote{\url{https://github.com/laiguokun/multivariate-time-series-data}}, ETT\footnote{\url{https://github.com/zhouhaoyi/ETDataset}} (ETTh1 and ETTm2), Electricity\footnote{\url{https://archive.ics.uci.edu/ml/datasets/ElectricityLoadDiagrams20112014}}, METR-LA\footnote{\url{https://github.com/liyaguang/DCRNN}} and ILI\footnote{\url{https://gis.cdc.gov/grasp/fluview/fluportaldashboard.html}} datasets. 
The overall statistics of these datasets are summarized in Table \ref{table C.1}. 

To show the non-stationarity of the six datasets, we especially choose the Augmented Dick-Fuller (ADF) test statistic used in~\cite{DBLP:conf/nips/LiuWWL22,DBLP:conf/nips/LiuLWL23,DBLP:conf/nips/LiuCLHLXC23} as the metric to quantitatively measure the degree of distribution shift.
A larger ADF test statistic means a higher level of non-stationarity, \textit{i.e.}, more severe distribution shifts. 
Based on the ADF test results in Table \ref{table C.1}, the MTS datasets adopted in our experiments show a high degree of distribution shift.
Notably, since the ADF statistical test of Weather\footnote{\url{https://www.bgc-jena.mpg.de/wetter/}} (-26.661 in \cite{DBLP:conf/nips/LiuLWL23}) is much smaller than other datasets, indicating relative stationarity, it is excluded from our evaluation benchmarks. Additionally, we use the more non-stationary METR-LA from the same transportation domain to replace Traffic\footnote{\url{http://pems.dot.ca.gov}} (-15.046 in \cite{DBLP:conf/nips/LiuLWL23}).
More dataset details are shown in Appendix \ref{secc.1}. 
We follow~\cite{DBLP:conf/nips/WuXWL21,DBLP:conf/aaai/ZhouZPZLXZ21} to preprocess data by the z-score normalization, and split all the datasets into training, validation, and test sets by the ratio of 7:1:2.

\subsubsection{Baselines}
We compare \textit{JointPGM} with the following state-of-the-art models for time series forecasting, including MLP-based models: Koopa~\cite{DBLP:conf/nips/LiuLWL23} and DLinear\cite{DBLP:conf/aaai/ZengCZ023}; Transformer-based models: Stationary~\cite{DBLP:conf/nips/LiuWWL22}, PatchTST~\cite{DBLP:conf/iclr/NieNSK23}, FEDformer~\cite{DBLP:conf/icml/ZhouMWW0022}, Autoformer~\cite{DBLP:conf/nips/WuXWL21},
iTransformer~\cite{liu2024itransformer} and Crossformer~\cite{DBLP:conf/iclr/ZhangY23}, and GNN-based model: WaveForM~\cite{DBLP:conf/aaai/YangLWZPW23}. Notably, Koopa and Stationary are specifically designed to tackle non-stationary forecasting challenges in time series, while iTransformer, Crossformer, and WaveForM are tailored for MTS forecasting.
Besides, we further compare \textit{JointPGM} with three model-agnostic normalization-based methods, including SAN~\cite{DBLP:conf/nips/LiuCLHLXC23}, Dish-TS~\cite{DBLP:conf/aaai/0010WWWZF23}, and RevIN~\cite{DBLP:conf/iclr/KimKTPCC22}, which respectively use Autoformer and FEDformer as backbones for non-stationary forecasting. 
More baseline details are provided in Appendix \ref{secc.2}. 
Regarding the evaluation metrics, we evaluate MTS forecasting performance using mean absolute error (MAE) and mean squared error (MSE). A lower MAE/MSE indicates better forecasting performance.
Each experiment is repeated three times with different seeds for each model on each dataset, and the mean of the test results is reported.

\subsubsection{Implementation Details}
All the experiments are implemented with Pytorch on an NVIDIA RTX 4090 24GB GPU. In our experiments, all mean functions $f_t^{\mu}(\cdot)$, $f_{\hat{z}}^{\mu}(\cdot)$, $f_{\tilde{Z}}^{\mu}(\cdot)$, $f_{\hat{t}}^{\mu}(\cdot)$, and variance functions $f_t^{\sigma^2}(\cdot)$, $f_{\hat{z}}^{\sigma^2}(\cdot)$, $f_{\tilde{Z}}^{\sigma^2}(\cdot)$, $f_{\hat{t}}^{\sigma^2}(\cdot)$ are instantiated as single linear layer. The depth of propagation $K$ in the \textit{Multi-hop Propagation} is set to 2 which is consistent with~\cite{DBLP:conf/kdd/WuPL0CZ20,DBLP:conf/aaai/YangLWZPW23} and the temperature parameter $\tau$ in the \textit{Gumbel-softmax Sampling} is set to 0.5 for all datasets. 
We denote the latent dimension size as $d$ and set $d$ to 128. 
In training, our model is trained using Adam optimizer with a learning rate of 1e-3, and the batch size is set to 128 for all datasets. 


\begin{figure*}[!t] 
\centering
\hspace{-0.37cm}
\subfloat[$L$ on ETTh1]{
    \label{fig4_1}
    \includegraphics[height=3.11cm]{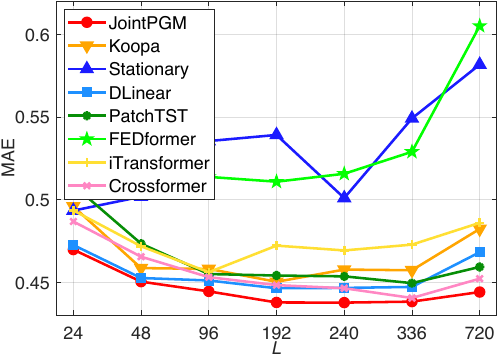}
    }
\hspace{-0.37cm}
\subfloat[$H$ on ETTh1]{
    \label{fig4_2}
    \includegraphics[height=3.11cm]{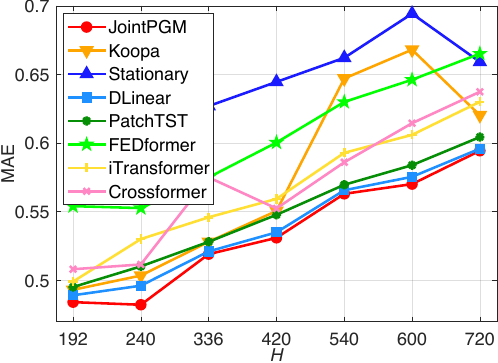}
    }
\hspace{-0.37cm}
\subfloat[$\alpha$ on Exchange]{
    \label{fig4_3}
    \includegraphics[height=3.11cm]{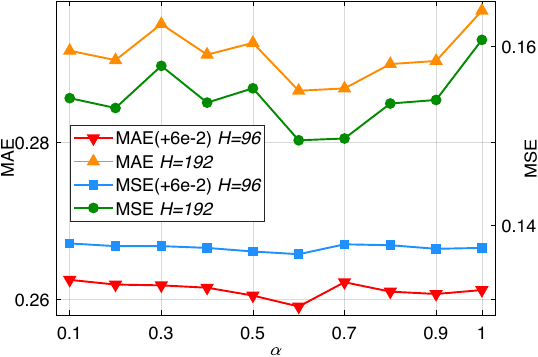}
    }
\hspace{-0.37cm}
\subfloat[$K$ on Exchange]{
    \label{fig4_4}
    \includegraphics[height=3.11cm]{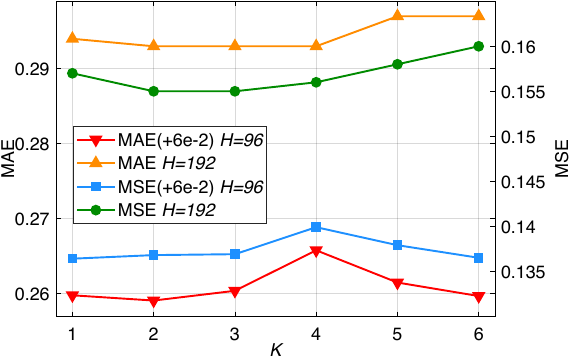}
    }
\hspace{-0.37cm}
\centering
\caption{Evaluation on model performance with different lookback length $L$ with the range ($24 \sim 720$), horizon length $H$ with the range ($192 \sim 720$), trade-off parameter $\alpha$ ($0.1 \sim 1$) and depth of propagation $K$ ($1 \sim 6$).}
\label{figure 4}
\end{figure*}

\begin{table}[!t]
\setlength{\tabcolsep}{3.6pt}
 \centering
 \caption{Ablation study in terms of forecasting accuracy. The lookback/horizon length is set to $L/H=24/48$ for ILI and $L/H=96/192$ for the others.}
 \label{table 2} 
 \newcommand{\tabincell}[2]{\begin{tabular}{@{}#1@{}}#2\end{tabular}}
 \begin{tabular}{lcccccccc}
 \toprule
 Dataset & \multicolumn{2}{c}{Exchange} & \multicolumn{2}{c}{ETTh1} & \multicolumn{2}{c}{Electricity} & \multicolumn{2}{c}{ILI} \\
 \cmidrule(r){2-3} \cmidrule(r){4-5} \cmidrule(r){6-7} \cmidrule(r){8-9}   
 Metrics & MAE & MSE & MAE & MSE & MAE & MSE & MAE & MSE  \\ 
 \midrule
 w/o DI & 0.299 & 0.162 & 0.487 & 0.493 & 0.281 & 0.195 & 1.371 & 3.784 \\
 w/o OF & 0.295 & 0.157 & 0.486 & 0.495 & 0.282 & 0.196 & 1.352 & 3.698 \\
 w/o ISE (A) & 0.426 & 0.331 & 0.589 & 0.620 & 0.357 & 0.257 & 1.402 & 3.998 \\ 
 w/o ISE (F) & 0.492 & 0.458 & 0.672 & 0.793 & 0.337 & 0.226 & 1.375 & 3.825 \\ 
 w/o IL & 0.295 & 0.158 & 0.486 & 0.492 & 0.281 & 0.195 & 1.365 & 3.761 \\
 w/o TG & 0.294 & 0.158 & 0.487 & 0.493 & 0.283 & 0.196 & 1.351 & 3.724 \\
 \textbf{JointPGM} & \textbf{0.293} & \textbf{0.155} & \textbf{0.484} & \textbf{0.491} & \textbf{0.279} & \textbf{0.187} & \textbf{1.340} & \textbf{3.640} \\
 \bottomrule 
 \end{tabular}
\end{table}

\subsection{Overall Performance}
Table \ref{table 1} showcases the forecasting results of \textit{JointPGM} compared to nine representative baselines with the best in \textbf{bold} and the second \underline{underlined}. 
From the table, we can observe \textit{JointPGM} achieves state-of-the-art performance in nearly \textbf{80$\%$} forecasting results with various prediction lengths.
Concretely, \textit{JointPGM} outperforms all general deep forecasting models across all time series datasets, with particularly notable improvements observed on datasets characterized by high non-stationarity: compared to their state-of-the-art results, we achieve 13.2$\%$ MSE reduction ($0.086 \rightarrow 0.076$) on Exchange (ADF: -1.902) and 4.8$\%$ ($4.000 \rightarrow 3.818$) on ILI (ADF: -5.334) under the horizon window of 96 and 24 respectively, which indicates that the potential of deep forecasting models is still constrained on non-stationary data.
Also, \textit{JointPGM} outperforms almost all deep models specifically designed to address distribution shifts. 
Notably, \textit{JointPGM} surpasses Stationary, the non-stationary version of Transformer, by a large margin, indicating that the traditional covariate shift assumption may not be consistent with the true distribution shift. This highlights the challenges posed by the diverse transitional shift patterns underlying the time series for model capacity.
Besides, different from Koopa disentangling series dynamics into time-variant dynamics and time-invariant dynamics using Koopman operators, \textit{JointPGM} achieves an average reduction of 2.7$\%$ in MAE and 9.3$\%$ in MSE by innovatively rethinking time-variant dynamics from both intra- and inter-series perspectives at a finer granularity.

\subsection{Comparison with Normalization-based Methods}
We further compare our performance with the advanced normalization-based methods including SAN, Dish-TS, and RevIN for addressing distribution shift. Table \ref{table c.2} has presented a performance comparison in MTS forecasting using Autoformer and FEDformer as backbones. From the results, we can observe that \textit{JointPGM} achieves the best performance in nearly \textbf{92$\%$} forecasting results compared to the existing normalization-based methods.
We attribute this superiority to the fine-grained capture of time-variant dynamics through explicit modeling of the time-variant transitional distribution between observations and predictions, while simultaneously considering the intra- and inter-series relationships inherent in the observations. This is demonstrated by the performance on two typical non-stationary datasets Exchange and ILI.
Specifically, compared to the second-best FEDformer+SAN, \textit{JointPGM} achieves an MSE reduction of 5.3$\%$ on the Exchange dataset and 6.6$\%$ on the ILI dataset. 
Notably, as shown in Table \ref{table 1} and Table \ref{table c.2}, \textit{JointPGM} exhibits slightly worse MAE than other compared models.
A potential explanation is that most compared models use a single MSE as an objective function while \textit{JointPGM} employs two MSE losses for prediction and reconstruction. Therefore, \textit{JointPGM} tends to prioritize improvements in MSE, consistently ranking first among all compared models except when $L/H=24/24$.

\subsection{Model Analysis}

\subsubsection{Hyperparameter Analysis}
To explore the impact of lookback length $L$, horizon length $H$, trade-off parameter $\alpha$, and the depth of propagation $K$, we conduct the following experiments for the sensitivity of these hyperparameters.

First, we investigate the impact of lookback length $L$ on the performance of the top-8 forecasting models on the ETTh1 dataset. In principle, extending the lookback window increases historical information availability, which will potentially improve forecasting performance. However, Figure \ref{fig4_1} demonstrates that Stationary and FEDformer, with Transformer-based architectures, have not benefited from a longer lookback window, aligning with the analysis in~\cite{DBLP:conf/aaai/ZhouZPZLXZ21}. Conversely, the remaining models consistently decrease MAE scores as the lookback window increases. Notably, our \textit{JointPGM} can capture the dynamics of intra- and inter-series correlations from more historical information to infer the time-variant transitional distribution in a fine-grained manner, thereby enhancing the prediction performance.
  

Then, we aim to discuss the influence of larger horizons (known as long-time series forecasting~\cite{DBLP:conf/aaai/ZhouZPZLXZ21,DBLP:conf/icde/LiLXTSJD23}) on the model performance. As Figure \ref{fig4_2} shows, when prolonging the horizon window to 720, \textit{JointPGM} consistently achieves superior forecasting performance compared to other baseline models. An intuitive reason is that larger horizons encompass more complex distribution changes and thus need more refined time-variant transitional distribution decomposition and modeling.

Furthermore, we study the impact of trade-off parameter $\alpha$ under the setting of $L=96, H=\{96,192\}$ on model performance on the Exchange dataset. Figure \ref{fig4_3} shows the performance comparison with different ratios $\alpha$ in Eq. (\ref{eqn17}).
We can observe that when $\alpha$ is less than 0.6, MAE and MSE all display a trend of violent fluctuation. Optimal performance is achieved with a moderate value for $\alpha$, resulting in the lowest MAE and MSE. A similar trend exists as increasing $\alpha$ from 0.6 since the unsuitable ratio of intra- and inter-series dynamics fails to fully describe the underlying causes of the distribution shift in series data and confuse latent data regularities. 

Lastly, we further examine how the depth of propagation $K$ affects the forecasting performance of \textit{JointPGM}. Figure \ref{fig4_4} has reported the experimental results on the Exchange dataset under the setting of $L=96,H=\{96,192\}$. From the results, we can draw the following conclusions: 1) Our \textit{JointPGM} outperforms the other compared models across different values of $K$, demonstrating the stability of our method; 2) Our \textit{JointPGM} achieves optimal performance when the value of $K$ is around 2, indicating that it is sufficient to propagate series information with 2 steps. As the depth of propagation $K$ increases, multi-hop propagation may suffer from over-smoothing caused by information aggregation, ultimately hindering forecasting performance.

\subsubsection{Ablation Study}
We perform an ablation study to assess the individual contributions of different key components in \textit{JointPGM} to the final performance. We list the MAE and MSE on four datasets under the setting of $L/H=24/48$ for ILI and $L/H=96/192$ for the others.
In particular, the following variants are examined: 1) \textbf{w/o DI}: Remove the entire dynamic inference from \textit{JointPGM} and exclude the loss term $\mathcal{L}_b$ from the overall loss function $\mathcal{L}$ (\textit{i.e.}, $\mathcal{L}_b=0$ in Eq. (\ref{eqn23})); 2) \textbf{w/o OF}: Replace the order-based time factors with timestamp-based time factors; 3) \textbf{w/o ISE (A)} and 4) \textbf{w/o ISE (F)}: Replace the entire independence-based series encoder with Autoformer and FEDformer encoders, and correspondingly substitute the decoder with their respective counterparts. Autoformer and FEDformer serve as frequently adopted backbones in normalization-based methods such as SAN~\cite{DBLP:conf/nips/LiuCLHLXC23}, Dish-TS~\cite{DBLP:conf/aaai/0010WWWZF23} and RevIN~\cite{DBLP:conf/iclr/KimKTPCC22}; 5) \textbf{w/o IL}: Completely remove the inter-series learner; 6) \textbf{w/o TG}: Remove the temporal gate from the intra-series learner. As the intra-series learner provides a crucial representation foundation for the inter-series learner, we remove this key component to validate its impact.

As presented in Table \ref{table 2}, the introduction of dynamic inference contributes to the forecasting performance, showing that comprehensively making the learned time factors more discriminative and sensitive to wild environments is vital for non-stationary forecasting models.
Additionally, we find it not easy to yield satisfactory results using timestamp-based time factors, which suggests the order-based time factor encoder may have successfully learned advantageous high-frequency patterns.
The incorporation of an independence-based series encoder significantly improves the model performance when compared with Autoformer- and FEDformer-based encoders, demonstrating the effectiveness of our strategy that jointly handles the distribution shift and models intra- and inter-series correlations.
The series encoder without the inter-series learner can provide performance improvements by focusing on modeling the intra-series distribution shift, but it still underperforms \textit{JointPGM} due to neglecting the shift caused by inter-series dynamics.
Furthermore, by equipping the intra-series learner with a temporal gate, \textit{JointPGM} can accurately capture time-variant dynamics within each series, further promoting the efficacy of non-stationary MTS forecasting.


\begin{figure}[!t]
\centering
\includegraphics[width=1\linewidth]{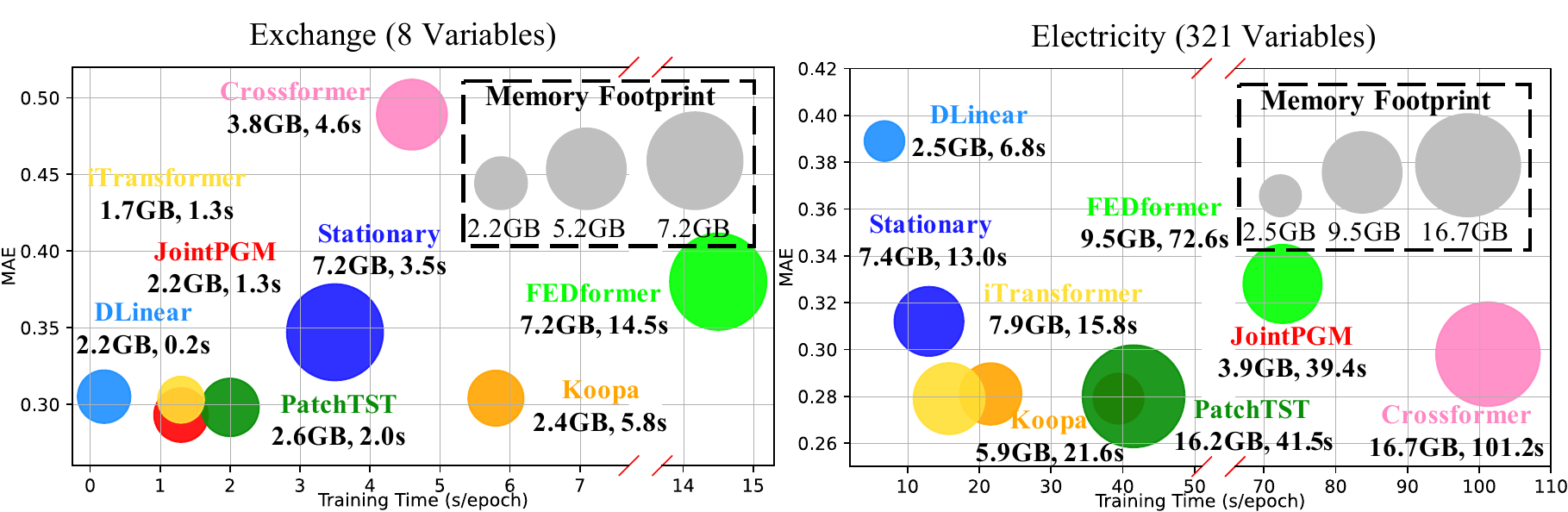}
\caption{Model efficiency comparison under the setting of $L/H=96/192$ of Exchange (left) and Electricity (right).}
\label{figure 5}
\end{figure}

\begin{figure}[!t] 
\centering
\hspace{-0.33cm}
\subfloat[w/o ISE (A)]{
    \label{fig11_1}
    \includegraphics[width=0.25\linewidth]{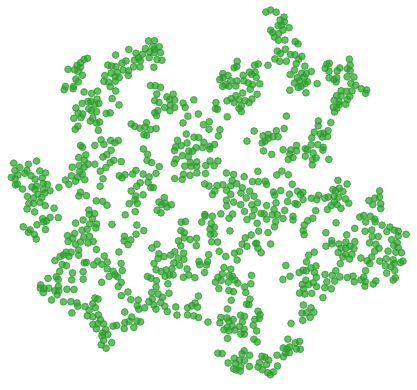}
    }
\hspace{-0.33cm}
\subfloat[w/o ISE (F)]{
    \label{fig11_2}
    \includegraphics[width=0.25\linewidth]{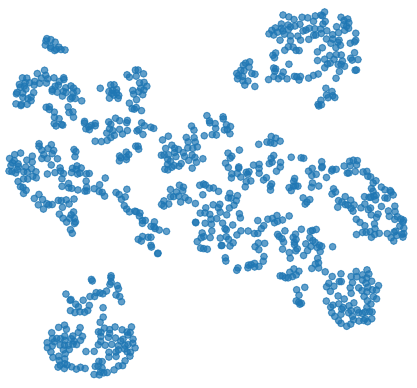}
    }
\hspace{-0.33cm}
\subfloat[w/o IL]{
    \label{fig11_3}
    \includegraphics[width=0.25\linewidth]{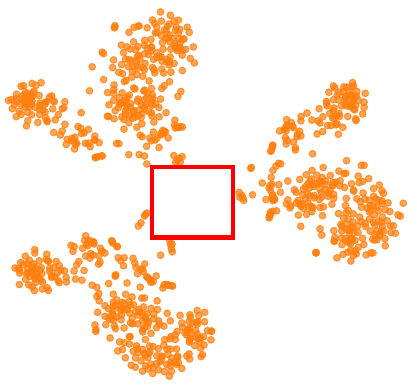}
    }
\hspace{-0.33cm}
\subfloat[JointPGM]{
    \label{fig11_4}
    \includegraphics[width=0.25\linewidth]{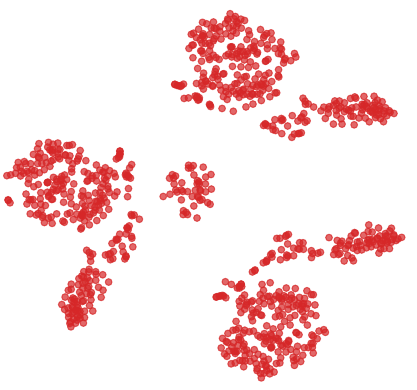}
    }
\hspace{-0.33cm}
\centering
\caption{t-SNE visualization of series representation $\tilde{\bm{H}}_{t-L:t}$ on the Electricity test set.}
\label{figure 11}
\end{figure}

\subsubsection{Model Efficiency}
We comprehensively evaluate the model efficiency of our \textit{JointPGM} and all baselines across three dimensions: forecasting performance, memory footprint, and training speed. Specifically, the forecasting performance (MAE) comes from Table \ref{table 1} under the setting of $L/H=96/192$. The memory footprint and training speed are calculated using the same batch size (128) and official code configuration. Figure \ref{figure 5} shows the efficiency results for two representative datasets of different scales: Exchange (8 variables, 7588 time steps) and Electricity (321 variables, 26304 time steps). The efficiency results on ETTm2 (7 Variables, 69680 Time Steps) are provided in Appendix \ref{secd.2}, Figure \ref{figure 12}.
From the figures, we can observe that: 1) In datasets with a relatively small number of variables (\textit{e.g.}, Exchange), the efficiency of \textit{JointPGM} is comparable to that of iTransformer, which only uses transformer encoders, and slightly inferior to that of the simplest linear model, DLinear.
For example, when compared to the MLP-based model Koopa customized for non-stationary forecasting, \textit{JointPGM} achieves a reduction of 77.6$\%$ in training time for the Exchange dataset, while maintaining a memory footprint of only 94.9$\%$;
2) In datasets with numerous variables (\textit{e.g.}, Electricity), the training speed is comparable to the SOTA forecasting model PatchTST, but \textit{JointPGM} has a significantly lower memory footprint. 
Meanwhile, \textit{JointPGM} achieves better forecasting performance than all other baselines in scenarios with numerous variables, as it is capable of jointly addressing distribution shifts and capturing the inherent intra- and inter-series correlations in MTS.

\subsection{Visualization Analysis}



\subsubsection{t-SNE Visualization of Series Representation}
To showcase the rationale behind studying fine-grained transitional shift from both intra- and inter-series perspectives, we visualize the feature distribution of series representation $\tilde{\bm{H}}_{t-L:t}$ using t-SNE for \textit{JointPGM} and its three variants \textbf{w/o ISE (A)}, \textbf{w/o ISE (F)} and \textbf{w/o IL}. For reliability, we randomly choose a batch of the Electricity test set and repeatedly run each experiment three times with different seeds. The overall results are depicted in Figure \ref{figure 11}.
The figure shows that: 1) The series representations learned from \textit{JointPGM} and the variant \textbf{w/o IL} exhibit a distinct clustering structure, indicating a robust and differentiated representation space. 
In contrast, those learned from the variants \textbf{w/o ISE (A)} and \textbf{w/o ISE (F)} reveal a more spread-out and less clustered pattern, suggesting that without the fine-grained decomposition, the series representations are comparatively less informative and distinguishable;
2) While the series representations learned from the variant \textbf{w/o IL} exhibit a clear clustering pattern by focusing only on intra-series transitional shift, they also show information loss (marked by the red box), possibly caused by misclustering due to spurious inter-series correlations. In contrast, those learned from \textit{JointPGM} are more evenly distributed across the entire 2D space. We further validate this rationale by showing heatmap visualizations of inter-series correlations in Appendix \ref{secd.4}, Figure \ref{figure 13}. Based on the insights from Figures \ref{figure 11} and \ref{figure 13}, \textit{JointPGM} can learn superior series representations to improve robustness against intricate distribution shifts and offer enhanced interpretability.

\subsubsection{Case Study of Forecasting}
\begin{figure}[!t]
\centering
\includegraphics[width=1\linewidth]{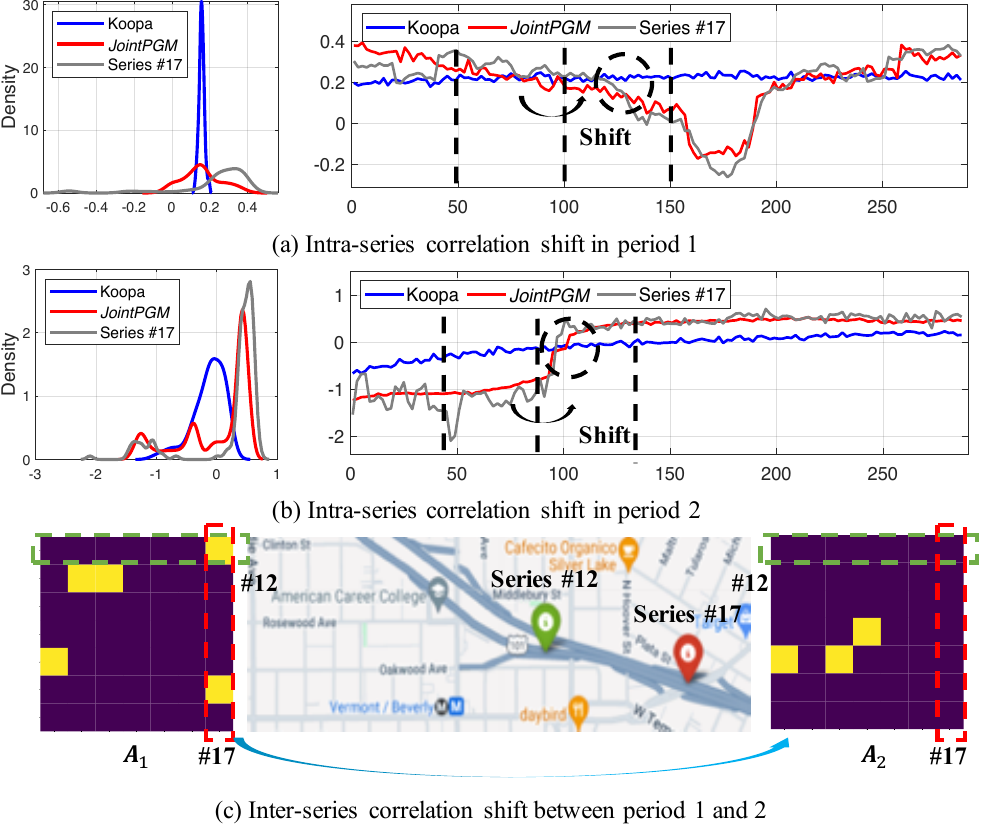}
\caption{Illustrative visualizations of intra- and inter-series correlation shifts on the METR-LA dataset.}
\vspace{-0.16cm}
\label{figure 7}
\end{figure}

We present a case study on real-world time series (METR-LA) in Figure \ref{figure 7}. We select one weekday (period 1) and one weekend (period 2) as the representative horizon windows. Firstly, we compare the predictions of series $\#$17 achieved by Koopa and our \textit{JointPGM} during these two periods at the data and distribution levels. We easily observe significant changes in series trend (could be regarded as significant changes in intra-series correlation, the black arrows), but Koopa cannot acquire accurate predictions. In contrast, our \textit{JointPGM} can perform precise predictions of future values and their distributions.  
The intuitive reason is when addressing the distribution shift, compared to coarse-grained Koopa, \textit{JointPGM} has jointly handled the distribution shift and modeled intra-/inter-series correlations, thereby boosting the performance.
Furthermore, we visualize the $\bm{A}_1$ and $\bm{A}_2$ learned by our \textit{JointPGM} during period 1 and 2. It can be observed that the learned correlation between series $\#$17 and other series also changes in different periods (the blue arrows), \textit{e.g.}, series $\#$17 exhibits a high correlation with series $\#$12 on weekday but not on weekends. This is reasonable since series $\#$12 is located near a school as indicated by the Google Map.
We present more forecasting showcases of \textit{JointPGM} and the three baselines: Koopa, Dish-TS, and RevIN, in Figures \ref{figure 8}, \ref{figure 9}, and \ref{figure 10} in Appendix \ref{secd.3}, respectively.

\section{Conclusion}
 This study aims to address the distribution shift problem to enhance the robustness of MTS forecasting by proposing a novel probabilistic graphical model and instantiating a neural framework, \textit{JointPGM}. Unlike previous normalization-based methods and time-variant models, \textit{JointPGM} deeply exploits the intrinsic causes of the distribution shift, boosting desirable model interpretability and the potential to enhance forecasting performance by jointly handling the distribution shift and modeling intra-/inter-series correlations. Experimentally, our model shows competitive performance on six real-world benchmarks with remarkable efficiency. Future works will explore time-variant dynamics on higher-dimensional MTS data and further improve efficiency.




\bibliographystyle{IEEEtran}
\bibliography{main_arxiv}

\vspace{11pt}


\vfill

\clearpage
\appendices

\section{Notations} \label{secA}
All symbols in our paper are carefully defined based on rules and we have provided detailed and clear explanations for the meanings of all symbols in Table \ref{table A.1}.

\begin{table}[ht]
  \caption{Notations used in the paper}
  \label{table A.1}
  \begin{tabular}{l|l}
    \toprule
    Symbol & Meaning \\
    \midrule
    $\bm{X}_{t-L:t}$ & \makecell[l]{the multivariate time series with a lookback window \\ of length-$L$ at time step $t$} \\
    $\bm{X}_{t:t+H}$ & \makecell[l]{the prediction target with a horizon window of length \\-$H$ at time step $t$} \\
    $\bm{x}_{t-L:t}^{(i)}$ & the $i^{th}$ series of $\bm{X}_{t-L:t}$  \\
    $\bm{x}_{t-L:t}^{(\overline{i})}$ & the complementary
    set of $\bm{x}_{t-L:t}^{(i)}$  \\
    $\bm{x}_{t:t+H}^{(i)}$ & the $i^{th}$ series of $\bm{X}_{t:t+H}$ \\
    $\bm{x}_{t:t+H}^{(\overline{i})}$ & the complementary
    set of $\bm{x}_{t:t+H}^{(i)}$ \\
    $N$ & the number of series \\
    $T$ & the number of time steps \\
    $\bm{H}_{t-L:t}$ & the representation of $\bm{X}_{t-L:t}$ \\
    $\bm{Z}_{t-L:t}$ & the latent variable of $\bm{X}_{t-L:t}$ \\
    $\hat{\bm{Z}}_{t-L:t}$ & the latent variable of $\bm{X}_{t-L:t}$ learned in intra-series learner \\
    $\tilde{\bm{Z}}_{t-L:t}$ & the latent variable of $\bm{X}_{t-L:t}$ learned in inter-series learner \\
    $\bm{h}_{t-L:t}^{(i)}$ & the representation of $\bm{x}_{t-L:t}^{(i)}$ \\
    $\bm{z}_{t-L:t}^{(i)}$ & the latent variable of $\bm{x}_{t-L:t}^{(i)}$ \\
    $\hat{\bm{z}}_{t-L:t}^{(i)}$ & the latent variable of $\bm{x}_{t-L:t}^{(i)}$ learned in intra-series learner \\
    $\tilde{\bm{z}}_{t-L:t}^{(i)}$ & the latent variable of $\bm{x}_{t-L:t}^{(i)}$ learned in inter-series learner \\
    $\bm{t}$ & the temporal order set \\
    $\bm{M}_t^{(1)}$ & the representation of $\bm{t}$ \\
    $\bm{Z}_t$ & the latent variable of $\bm{t}$ inferred from $\bm{t}$ \\
    $\hat{\bm{Z}}_t$ & the latent variable of $\bm{t}$ inferred from $\bm{Z}_{t-L:t}$ \\
    $\bm{G}$ & the temporal gate \\
    $\bm{W}_t$ & the continuous adjacency matrix at time step $t$ \\
    $\bm{A}_t$ & the discrete adjacency matrix at time step $t$ \\
    $\bm{\mu}_t$ & the mean at time step $t$ \\
    $\bm{\sigma}_t^2$ & the variance at time step $t$ \\
    $d$ & the latent dimension size \\
    $b$ & the Fourier feature size \\
    $\sigma_s$ & the scale hyperparameter and $s$ is the corresponding index \\
    $K$ & the depth of propagation in multi-hop propagation \\
    $\tau$ & the temperature parameter in Gumbel-softmax sampling \\
    $\alpha$ & the trade-off parameter \\
    $f^{\mu}(\cdot)$ & the multivariate function to learn the mean vector \\
    $f^{\sigma^2}(\cdot)$ & the multivariate function to learn the variance vector \\
    $\mathcal{P}(\cdot|\cdot)$ & the transitional/conditional distribution \\
    $\mathbb{KL}(\cdot|\cdot)$ & the Kullback–Leibler divergence \\
  \bottomrule
\end{tabular}
\end{table}

\section{The Details of Theoretical Analyses} \label{secB}

\subsection{The proof of Proposition 1} \label{secB.1}
Since our objective is for the learned time series and their corresponding time factors to exhibit strong discriminative characteristics and closely align with each other, we perform the following variational approximation with the Kullback-Leibler (KL) divergence:
\begin{equation} \label{eqn5}
    \begin{split}
    &\mathbb{KL}[\mathcal{P}_\psi(\bm{X}_H,\bm{Z}_L,\bm{Z}_t|\bm{X}_L)||\mathcal{P}_\phi(\bm{X}_H,\bm{Z}_L,\bm{Z}_t|\bm{X}_L,\bm{t})] \\
    &=\int_{\bm{X}_H}\int_{\bm{Z}_L}\int_{\bm{Z}_t}\mathcal{P}_\psi(\bm{X}_H,\bm{Z}_L,\bm{Z}_t|\bm{X}_L) \\ 
    &\qquad \qquad  \qquad \qquad  \log\frac{\mathcal{P}_\psi(\bm{X}_H,\bm{Z}_L,\bm{Z}_t|\bm{X}_L)}{\mathcal{P}_\phi(\bm{X}_H,\bm{Z}_L,\bm{Z}_t|\bm{X}_L,\bm{t})} \dif \bm{X}_H \dif \bm{Z}_L \dif \bm{Z}_t  \\
    &=\int_{\bm{X}_H}\int_{\bm{Z}_L}\int_{\bm{Z}_t}\mathcal{P}_\psi(\bm{X}_H|\bm{Z}_L)\mathcal{P}_\psi(\bm{Z}_L,\bm{Z}_t|\bm{X}_L) \log[\frac{\mathcal{P}_\psi(\bm{Z}_L,\bm{Z}_t|\bm{X}_L)}{\mathcal{P}_\phi(\bm{Z}_L,\bm{Z}_t|\bm{X}_L,\bm{t})} \cdot \\
    &\qquad \qquad  \qquad \qquad  \frac{\mathcal{P}_\psi(\bm{X}_H|\bm{Z}_L)}{\mathcal{P}_\phi(\bm{X}_H|\bm{X}_L,\bm{t})}] \dif \bm{X}_H \dif \bm{Z}_L \dif \bm{Z}_t  \\
    &=\int_{\bm{X}_H}\int_{\bm{Z}_L}\int_{\bm{Z}_t}\mathcal{P}_\psi(\bm{X}_H|\bm{Z}_L)\mathcal{P}_\psi(\bm{Z}_L,\bm{Z}_t|\bm{X}_L) \log\frac{\mathcal{P}_\psi(\bm{Z}_L,\bm{Z}_t|\bm{X}_L)}{\mathcal{P}_\phi(\bm{Z}_L,\bm{Z}_t|\bm{X}_L,\bm{t})} \\
    &\qquad \qquad  \qquad \qquad  \dif \bm{X}_H \dif \bm{Z}_L \dif \bm{Z}_t \\
    &\quad +\int_{\bm{X}_H}\int_{\bm{Z}_L}\int_{\bm{Z}_t}\mathcal{P}_\psi(\bm{X}_H|\bm{Z}_L)\mathcal{P}_\psi(\bm{Z}_L,\bm{Z}_t|\bm{X}_L) \log\frac{\mathcal{P}_\psi(\bm{X}_H|\bm{Z}_L)}{\mathcal{P}_\phi(\bm{X}_H|\bm{X}_L,\bm{t})} \\
    &\qquad \qquad  \qquad \qquad  \dif \bm{X}_H \dif \bm{Z}_L \dif \bm{Z}_t  \\
    &=\int_{\bm{X}_H}\mathcal{P}_\psi(\bm{X}_H|\bm{Z}_L)\mathbb{KL}[\mathcal{P}_\psi(\bm{Z}_L,\bm{Z}_t|\bm{X}_L) || \mathcal{P}_\phi(\bm{Z}_L,\bm{Z}_t|\bm{X}_L,\bm{t})] \dif \bm{X}_H  \\
    &\quad +\int_{\bm{Z}_L}\int_{\bm{Z}_t}\mathcal{P}_\psi(\bm{Z}_L,\bm{Z}_t|\bm{X}_L) \mathbb{KL}[\mathcal{P}_\psi(\bm{X}_H|\bm{Z}_L)||\mathcal{P}_\phi(\bm{X}_H|\bm{X}_L,\bm{t})] \\
    &\qquad \qquad  \qquad \qquad  \dif \bm{Z}_L \dif \bm{Z}_t \\
    &=\mathbb{KL}[\mathcal{P}_\psi(\bm{Z}_L,\bm{Z}_t|\bm{X}_L) || \mathcal{P}_\phi(\bm{Z}_L,\bm{Z}_t|\bm{X}_L,\bm{t})] \\
    &\quad +\mathbb{E}_{(\bm{Z}_L,\bm{Z}_t) \sim \mathcal{P}_\psi(\bm{Z}_L,\bm{Z}_t|\bm{X}_L)} \mathbb{KL}[\mathcal{P}_\psi(\bm{X}_H|\bm{Z}_L)||\mathcal{P}_\phi(\bm{X}_H|\bm{X}_L,\bm{t})],    
    \end{split}
\end{equation}

The first term in Eq. (\ref{eqn5}) is:
\begin{equation} \label{eqn6}
    \begin{split}
    &\mathbb{KL}[\mathcal{P}_\psi(\bm{Z}_L,\bm{Z}_t|\bm{X}_L) || \mathcal{P}_\phi(\bm{Z}_L,\bm{Z}_t|\bm{X}_L,\bm{t})] \\
    &=\int_{\bm{Z}_L}\int_{\bm{Z}_t}\mathcal{P}_\psi(\bm{Z}_L,\bm{Z}_t|\bm{X}_L)\log\frac{\mathcal{P}_\psi(\bm{Z}_L,\bm{Z}_t|\bm{X}_L)}{\mathcal{P}_\phi(\bm{Z}_L,\bm{Z}_t|\bm{X}_L,\bm{t})} \dif \bm{Z}_L \dif \bm{Z}_t  \\
    &=\int_{\bm{Z}_L}\int_{\bm{Z}_t}\mathcal{P}_\psi(\bm{Z}_L|\bm{X}_L)\mathcal{P}_\theta(\bm{Z}_t|\bm{Z}_L)\log\frac{\mathcal{P}_\psi(\bm{Z}_L|\bm{X}_L)\mathcal{P}_\theta(\bm{Z}_t|\bm{Z}_L)}{\mathcal{P}_\phi(\bm{Z}_L|\bm{X}_L,\bm{t})\mathcal{P}_\phi(\bm{Z}_t|\bm{t})} \\
    &\qquad \qquad  \qquad \qquad  \dif \bm{Z}_L \dif \bm{Z}_t  \\
    &=\int_{\bm{Z}_L}\int_{\bm{Z}_t}\mathcal{P}_\psi(\bm{Z}_L|\bm{X}_L)\mathcal{P}_\theta(\bm{Z}_t|\bm{Z}_L)\log\frac{\mathcal{P}_\psi(\bm{Z}_L|\bm{X}_L)}{\mathcal{P}_\phi(\bm{Z}_L|\bm{X}_L,\bm{t})} \dif \bm{Z}_L \dif \bm{Z}_t \\
    &\quad +\int_{\bm{Z}_L}\int_{\bm{Z}_t}\mathcal{P}_\psi(\bm{Z}_L|\bm{X}_L)\mathcal{P}_\theta(\bm{Z}_t|\bm{Z}_L)\log\frac{\mathcal{P}_\theta(\bm{Z}_t|\bm{Z}_L)}{\mathcal{P}_\phi(\bm{Z}_t|\bm{t})} \dif \bm{Z}_L \dif \bm{Z}_t \\
    &=\int_{\bm{Z}_t}\mathcal{P}_\theta(\bm{Z}_t|\bm{Z}_L) \mathbb{KL}[\mathcal{P}_\psi(\bm{Z}_L|\bm{X}_L) || \mathcal{P}_\phi(\bm{Z}_L|\bm{X}_L,\bm{t})] \dif \bm{Z}_t \\
    &\quad +\int_{\bm{Z}_L}\mathcal{P}_\psi(\bm{Z}_L|\bm{X}_L) \mathbb{KL}[\mathcal{P}_\theta(\bm{Z}_t|\bm{Z}_L) || \mathcal{P}_\phi(\bm{Z}_t|\bm{t})] \dif \bm{Z}_L  \\ 
    &=\mathbb{KL}[\mathcal{P}_\psi(\bm{Z}_L|\bm{X}_L) || \mathcal{P}_\phi(\bm{Z}_L|\bm{X}_L,\bm{t})] \\
    &\quad +\mathbb{E}_{\bm{Z}_L \sim \mathcal{P}_\psi(\bm{Z}_L|\bm{X}_L)} \mathbb{KL}[\mathcal{P}_\theta(\bm{Z}_t|\bm{Z}_L) || \mathcal{P}_\phi(\bm{Z}_t|\bm{t})],
    \end{split}
\end{equation}
Combining Eq. (\ref{eqn5}) and Eq. (\ref{eqn6}), we complete the proof.

Note that the latent variables $\bm{Z}_L$ and $\bm{Z}_t$ are \textit{not inter-dependent}. If $\bm{Z}_L$ and $\bm{Z}_t$ are inter-dependent, the inference of \textit{$\bm{Z}_L$ (resp. $\bm{Z}_t$) must necessitate $\bm{Z}_t$ (resp. $\bm{Z}_L$)}. Therefore, in Eq. (\ref{eqn6}), there should be $\mathcal{P}_\psi(\bm{Z}_L|\bm{X}_L,\bm{Z}_t)$ and $\mathcal{P}_\phi(\bm{Z}_t|\bm{t},\bm{Z}_L)$ rather than $\mathcal{P}_\psi(\bm{Z}_L|\bm{X}_L)$ and $\mathcal{P}_\phi(\bm{Z}_t|\bm{t})$. As Figure \ref{figure 2} shows, while $\bm{Z}_L$ are jointly inferred by $\bm{X}_L$ and $\bm{t}$, $\bm{Z}_t$ can be inferred without $\bm{X}_L$ but using $\bm{t}$. Therefore, $\mathcal{P}_\psi(\bm{Z}_L|\bm{X}_L,\bm{t})$ and $\mathcal{P}_\phi(\bm{Z}_t|\bm{t})$ hold in Eq. (\ref{eqn6}).

\subsection{The derivation of Eq. (\ref{eqn21})} \label{secB.2} 
For term ($a$), \textit{i.e.}, $\mathbb{KL}[\mathcal{P}_{\psi}(\bm{Z}_L|\bm{X}_L)||\mathcal{P}_{\phi}(\bm{Z}_L|\bm{X}_L,\bm{t})]$, it aims to keep the inference using the variational distribution and the inference using the posterior is \textit{close}, which also guarantees the reliable and high-quality sampling. We employ the variational evidence lower bound (ELBO) to constrain the term ($a$). Suppose that $\mathcal{P}(\bm{X}_L)$ is a constant given $\bm{X}_L$. It is equivalent to maximizing the ELBO and minimizing $\mathcal{L}_a$. Mathematically, we have: 
\begin{equation} \label{eqn22}
    \begin{split}
    \log&\mathcal{P}(\bm{X}_L)=\mathbb{E}_{\bm{Z}_L \sim \mathcal{P}_{\psi}(\bm{Z}_L|\bm{X}_L)}[\log\mathcal{P}(\bm{X}_L)] \\
    &=\mathbb{E}_{\bm{Z}_L \sim \mathcal{P}_{\psi}(\bm{Z}_L|\bm{X}_L)}[\log\frac{\mathcal{P}(\bm{X}_L|\bm{Z}_L)\mathcal{P}(\bm{Z}_L)}{\mathcal{P}_{\phi}(\bm{Z}_L|\bm{X}_L,\bm{t})}] \\
    &=\mathbb{E}_{\bm{Z}_L \sim \mathcal{P}_{\psi}(\bm{Z}_L|\bm{X}_L)}[\log\frac{\mathcal{P}(\bm{X}_L|\bm{Z}_L)\mathcal{P}(\bm{Z}_L)}{\mathcal{P}_{\phi}(\bm{Z}_L|\bm{X}_L,\bm{t})} \cdot \frac{\mathcal{P}_{\psi}(\bm{Z}_L|\bm{X}_L)}{\mathcal{P}_{\psi}(\bm{Z}_L|\bm{X}_L)}] \\
    &=\mathbb{E}_{\bm{Z}_L}[\log\mathcal{P}(\bm{X}_L|\bm{Z}_L)]-\mathbb{E}_{\bm{Z}_L}[\log\frac{\mathcal{P}_{\psi}(\bm{Z}_L|\bm{X}_L)}{\mathcal{P}(\bm{Z}_L)}]  \\
    &\quad +\mathbb{E}_{\bm{Z}_L}[\log\frac{\mathcal{P}_{\psi}(\bm{Z}_L|\bm{X}_L)}{\mathcal{P}_{\phi}(\bm{Z}_L|\bm{X}_L,\bm{t})}] \\
    &=\mathbb{E}_{\bm{Z}_L}[\log\mathcal{P}(\bm{X}_L|\bm{Z}_L)]-\mathbb{KL}[\mathcal{P}_{\psi}(\bm{Z}_L|\bm{X}_L)||\mathcal{P}(\bm{Z}_L)] \\
    &\quad +\mathbb{KL}[\mathcal{P}_{\psi}(\bm{Z}_L|\bm{X}_L)||\mathcal{P}_{\phi}(\bm{Z}_L|\bm{X}_L,\bm{t})] \\
    &=\underbrace{\mathbb{E}_{\bm{Z}_L}[\log\mathcal{P}(\bm{X}_L|\bm{Z}_L)]-\mathbb{KL}[\mathcal{P}_{\psi}(\bm{Z}_L|\bm{X}_L)||\mathcal{P}(\bm{Z}_L)]}_{ELBO}+\mathcal{L}_{a},
    \end{split}
\end{equation}
Thus:
\begin{equation} \label{eqn24}
    \begin{split}
    \mathcal{L}_a&=-ELBO \\
    &=-\mathbb{E}_{\bm{Z}_L}[\log\mathcal{P}(\bm{X}_L|\bm{Z}_L)]+\mathbb{KL}[\mathcal{P}_{\psi}(\bm{Z}_L|\bm{X}_L)||\mathcal{P}(\bm{Z}_L)] \\ 
    &=l(\hat{\bm{X}}_L-\bm{X}_L) + (-\log\bm{\sigma}_Z+\frac{1}{2}\bm{\sigma}^2_Z+\frac{1}{2}\bm{\mu}_Z^2-\frac{1}{2}),
    \end{split}
\end{equation}
where $l$ denotes a distance metric for which we use the MSE loss,
$\bm{\mu}_{Z}$ and $\bm{\sigma}^2_Z$ are the mean and variance vectors of $\bm{Z}_L$.

\section{Deep Discussions of Our Method \textit{JointPGM}} \label{sece}

\subsection{Why can our proposed \textit{JointPGM} address the non-stationary MTS forecasting problem?}
As stated in the introduction, we argue there are two primary categories of approaches to address the non-stationary time series forecasting issue. The first (e.g., RevIN\cite{DBLP:conf/iclr/KimKTPCC22}) is to alleviate temporal mean and covariance shift by normalization, and the second (e.g., Koopa\cite{DBLP:conf/nips/LiuLWL23}) is to learn time-variant series dynamics. Our \textit{JointPGM} belongs to the second category. Different from Koopa\cite{DBLP:conf/nips/LiuLWL23} disentangling the series into time-variant and time-invariant dynamics, we rethink the time-variant dynamics from the intra- and inter-series perspective from a finer granularity. Since non-stationarity is intrinsically attributed to distributional shift, learning time-variant dynamics essentially stems from modeling the time-variant transitional distribution between inputs and outputs. Hence, we decompose such time-variant transitional distribution into intra- and inter-series parts. As shown in Figure \ref{fig2_3}, we introduce dynamic time factors $\bm{t}$ as conditions to regulate this process and explicitly model the intra-series transitional distribution $\mathcal{P}(\bm{x}_{t:t+H}^{(i)}|\bm{x}_{t-L:t}^{(i)}, \bm{t})$ and inter-series transitional distribution $\mathcal{P}(\bm{x}_{t:t+H}^{(i)}|\bm{x}_{t-L:t}^{(\Bar{i})}, \bm{t})$, see Section \ref{sec3.B.2}. By this means, time-variant dynamics are finely learned, thereby addressing the non-stationary time series forecasting issues.

\subsection{Why do we need a probabilistic graphical model?}
A probabilistic graphical model (PGM) can represent a probability distribution of random variables and provide a principled and interpretable manner for exploiting dependency relationships among variables. Inspired by this, we represent and decompose time-variant transitional distribution and instantiate a neural network \textit{JointPGM} based on the PGM. It can reveal the intrinsic causes of the transitional distribution, ensuring desirable model interpretability and the potential to enhance forecasting performance by jointly addressing the distribution shift and capturing intra-/inter-series correlations. The ablation study (see Table \ref{table 2}), series representation visualization (see Figures \ref{figure 11} and \ref{figure 13}), and case study of forecasting (see Figure \ref{figure 7}) all explain the rationality behind this decomposition.

\subsection{How is the interpretability of our \textit{JointPGM} demonstrated?}
We highlight that our interpretability is rooted in the method of decomposition from intra- and inter-series dimensions using the probabilistic graphical model (PGM).

\textit{In the method}, PGM can represent the probability distribution of random variables and provide an interpretable manner to explore dependency relationships among variables. We present a tailored and specific PGM framework for decomposing time-variant transitional distribution, and accordingly instantiate the proposed neural network \textit{JointPGM} based on the PGM framework (see Section \ref{sec3.2}), ensuring desirable model interpretability in terms of understanding the distribution shift. Moreover, compared with the previous MSE loss function, the decomposition of our optimization objective is also interpretable. It can perform different decomposition terms as constraints for different sub-procedures in PGM to help the understanding of what role they play in complex non-stationary forecasting, \textit{e.g.}, term (c) can ensure the generation from raw data and latent variables is consistent.

\textit{In the experiments}, we have presented series representation visualization (see Figures \ref{figure 11} and \ref{figure 13}), conducted the case study of forecasting on the METR-LA dataset (see Figure \ref{figure 7}), and compared to different variants like w/o ISE (A) in ablation study (see Table \ref{table 2}). These experiments strongly support the rationality of this decomposition method.

\subsection{How do we avoid the potential overfitting risk with the introduction of many latent variables?}
We have adopted effective strategies to mitigate potential overfitting risk, \textit{e.g.}, we instantiated the mean and variance functions of Gaussian sampling as a single linear layer, effectively reducing the parameters and training time. We also carefully tuned the latent dimension size $d$ in $\{64,...,512\}$ and selected the optimal size of 128 to avoid inferior performance due to possible overfitting. The superior forecasting performance and efficiency on small-scale datasets like ETTh1 demonstrate the effectiveness in mitigating overfitting.

\begin{table*}[!t] 
\setlength{\tabcolsep}{0.5pt}
 \centering
 \caption{Performance comparison to Dish-TS~\cite{DBLP:conf/aaai/0010WWWZF23} and RevIN~\cite{DBLP:conf/iclr/KimKTPCC22} with different backbones including Transformer (T), Informer (I) and Autoformer (A). The Lookback/horizon length is set to $L/H=24/48$ for the ILI and $L/H=96/96$ for the others. Different from Table \ref{table c.2}, the results are returned to their original values according to the settings in reference~\cite{DBLP:conf/aaai/0010WWWZF23}. \textbf{Bold} indicates the best result, while \underline{underlining} indicates the second-best result.}
 \label{table C.3}
 \newcommand{\tabincell}[2]{\begin{tabular}{@{}#1@{}}#2\end{tabular}}
 \begin{tabular}{c|cccccccccccc}
  \toprule
  Datasets & \multicolumn{2}{c}{Exchange} & \multicolumn{2}{c}{ETTm2} & \multicolumn{2}{c}{ETTh1} & \multicolumn{2}{c}{Electricity} & \multicolumn{2}{c}{METR-LA} & \multicolumn{2}{c}{ILI}   \\
  \cmidrule(r){2-3} \cmidrule(r){4-5} \cmidrule(r){6-7} \cmidrule(r){8-9} \cmidrule(r){10-11} \cmidrule(r){12-13} 
  Metrics & MAE($\times e^{-1}$) & MSE($\times e^{-1}$) & MAE($\times e^1$) & MSE($\times e^2$) & MAE($\times e^1$) & MSE($\times e^2$) & MAE($\times e^3$) & MSE($\times e^8$) & MAE($\times e^2$) & MSE($\times e^3$) & MAE($\times e^5$) & MSE($\times e^{11}$)   \\ 
  \midrule 
  Dish-TS+T & 0.249 & 0.014 & 0.205 & 0.105 & 0.192 & 0.119 & 0.299 & 0.079 & 0.119 & 0.451 & 0.303 & 0.079 \\
  Dish-TS+I & 0.674 & 0.149 & 0.227 & 0.125 & 0.212 & 0.140 & 0.339 & 0.106 & \textbf{0.114} & 0.416 & 0.301 & 0.073 \\
  Dish-TS+A & 0.278 & 0.018 & 0.217 & 0.112 & 0.241 & 0.195 & 0.307 & 0.082 & 0.156 & 0.535 & 0.288 & \textbf{0.072} \\
  \midrule 
  RevIN+T & 0.253 & 0.015 & 0.234 & 0.133 & 0.348 & 0.327 & 0.975 & 0.771 & 0.134 & 0.495 & 0.402 & 0.124 \\
  RevIN+I & 0.241 & 0.015 & 0.223 & 0.126 & 0.208 & 0.141 & 0.370 & 0.130 & 0.128 & 0.451 & \textbf{0.286} & 0.076 \\
  RevIN+A & 0.276 & 0.018 & 0.219 & 0.118 & 0.197 & 0.121 & 0.293 & 0.070 & 0.144 & 0.524 & 0.300 & 0.079 \\
  \midrule 
  \textbf{JointPGM} & \textbf{0.176} & \textbf{0.008} & \textbf{0.198} & \textbf{0.099} & \textbf{0.178} & \textbf{0.111} & \textbf{0.238} & \textbf{0.057} & 0.120 & \textbf{0.370} & 0.592 & 0.266  \\
  \bottomrule 
 \end{tabular}
\end{table*}

\begin{figure}[!t] 
\centering
\includegraphics[width=1\linewidth]{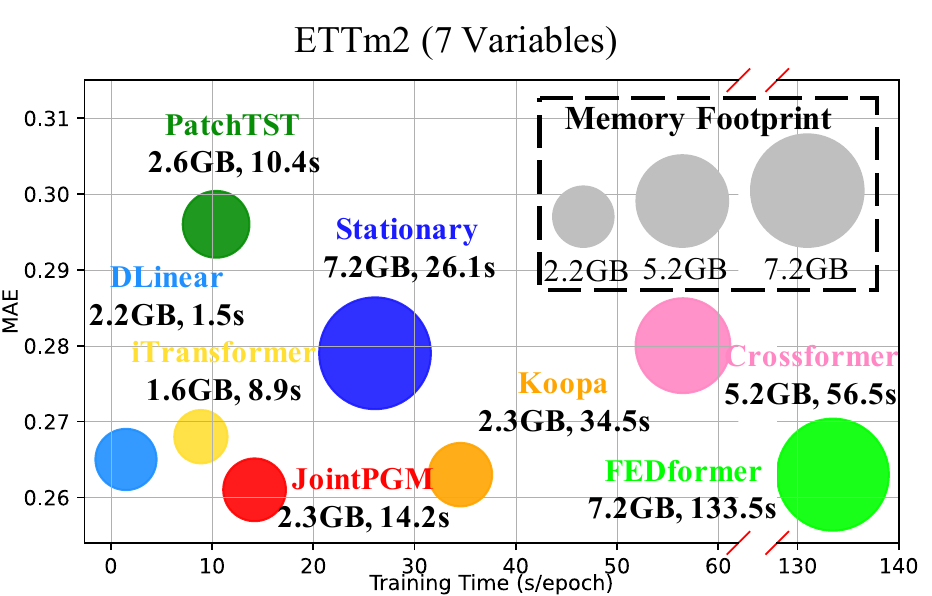}
\caption{Model efficiency comparison under the setting of $L/H=96/192$ on ETTm2 (7 Variables, 69680 Time Steps).}
\label{figure 12}
\end{figure}

\section{More Experimental Details}   \label{secc}

\subsection{More dataset details}  \label{secc.1}
We adopt six real-world benchmarks in the experiments to evaluate the non-stationary MTS forecasting task. The overall statistics of these datasets are summarized in Table \ref{table C.1}.
The experimental results (see Table \ref{table 1}) on these datasets are better than those of baselines, which sufficiently proves that our proposed method is exactly superior and effective in handling the distribution shift problem in MTS forecasting.

The details of these datasets are as follows:
\begin{itemize}
    \item \textbf{Exchange:\footnote{\url{https://github.com/laiguokun/multivariate-time-series-data}}}~comprises daily exchange rates of eight foreign countries, namely Australia, Britain, Canada, Switzerland, China, Japan, New Zealand, and Singapore, spanning from 1990 to 2016, with a sampling frequency of 1 day.
    \item \textbf{ETT:\footnote{\url{https://github.com/zhouhaoyi/ETDataset}}}~is sourced from two distinct electric transformers labeled as 1 and 2, each offering two different resolutions: 15-minute (denoted as `m') and 1-hour (denoted as `h'). We designate \textbf{ETTh1} and \textbf{ETTm2} as our benchmarks for non-stationary time series forecasting. 
    \item \textbf{Electricity:\footnote{\url{https://archive.ics.uci.edu/ml/datasets/ElectricityLoadDiagrams20112014}}}~comprises the electricity consumption of 321 clients for MTS forecasting, collected since 01/01/2011 with a sampling frequency of every 15 minutes.
    \item \textbf{METR-LA:\footnote{\url{https://github.com/liyaguang/DCRNN}}}~contains traffic information gathered from loop detectors on the highways of Los Angeles Country. It comprises data from 207 sensors spanning from 01/03/2012 to 30/06/2012 with a sampling frequency of every 5 minutes.
    \item \textbf{ILI:\footnote{\url{https://gis.cdc.gov/grasp/fluview/fluportaldashboard.html}}}~contains the weekly recorded data on patients with influenza-like illness (ILI) from the Centers for Disease Control and Prevention of the United States spanning from 2002 to 2021. This data describes the ratio of patients seen and the total number of patients. 
\end{itemize}



\begin{figure*}[!t] 
\centering
    \includegraphics[height=0.258\textheight]{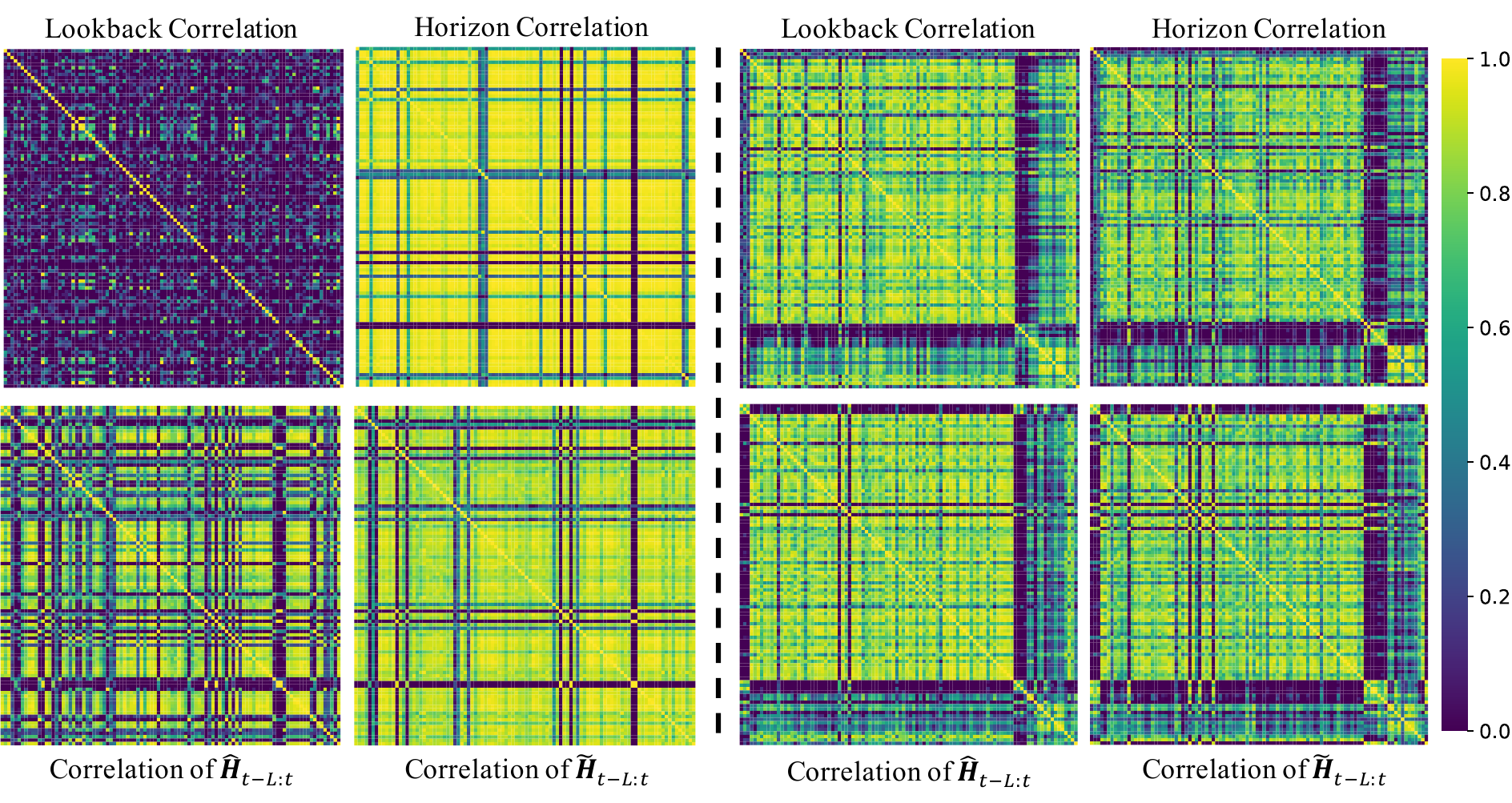}
\caption{Inter-series correlation visualization of $\bm{X}_{t-L:t}$, $\hat{\bm{H}}_{t-L:t}$, $\tilde{\bm{H}}_{t-L:t}$ and $\bm{X}_{t:t+H}$. The left of the dashed line is the METR-LA test set, and the right is the Electricity test set.}
\label{figure 13}
\end{figure*}

\subsection{More baseline details}  \label{secc.2}
We adopt two representative non-stationary forecasting models including Koopa~\cite{DBLP:conf/nips/LiuLWL23} and Stationary~\cite{DBLP:conf/nips/LiuWWL22} and seven state-of-the-art general deep forecasting models including DLinear\cite{DBLP:conf/aaai/ZengCZ023}, 
PatchTST~\cite{DBLP:conf/iclr/NieNSK23}, 
FEDformer~\cite{DBLP:conf/icml/ZhouMWW0022}, Autoformer~\cite{DBLP:conf/nips/WuXWL21}, iTransformer~\cite{liu2024itransformer}, Crossformer~\cite{DBLP:conf/iclr/ZhangY23}, and WaveForM~\cite{DBLP:conf/aaai/YangLWZPW23} for comparison. Additionally, we further compare \textit{JointPGM} with three model-agnostic normalization-based methods SAN~\cite{DBLP:conf/nips/LiuCLHLXC23}, Dish-TS~\cite{DBLP:conf/aaai/0010WWWZF23} and RevIN~\cite{DBLP:conf/iclr/KimKTPCC22} with different backbones for non-stationary forecasting. 
Note that although Stationary is based on normalization, we classify it under non-stationary models for discussion due to its exclusive focus on the Transformer architecture.
We introduce these models as follows:

\begin{itemize}
    \item \textbf{Koopa} uses stackable blocks to learn hierarchical dynamics, Koopman operators for transition modeling, and context-aware operators for handling time-variant dynamics. We implement their provided source code from: \href{https://github.com/thuml/Koopa}{https://github.com/thuml/Koopa}. 
    \item \textbf{Stationary} introduces a series stationarization module with statistics for better predictability and a de-stationary attention module to re-integrate the inherent non-stationary information for non-stationary MTS forecasting. We implement their provided source code from: \href{https://github.com/thuml/Nonstationary_Transformers}{https://github.com/thuml/Nonstationary\_Transformers}. 
    \item \textbf{DLinear}: decomposes series data into trend and seasonal components using a moving average kernel and applies linear layers to each component. We implement their provided source code from: \href{https://github.com/honeywell21/DLinear}{https://github.com/honeywell21/DLinear}.
    \item \textbf{PatchTST}: is a Transformer-based model for time series forecasting tasks by introducing two key components: patching and channel-independent structure. We implement their provided source code from: \href{https://github.com/PatchTST}{https://github.com/PatchTST}. 
    \item \textbf{FEDformer}: combines an attention mechanism incorporating low-rank approximation in frequency with a mixture of expert decomposition to manage distribution shifting. We implement their provided source code from: \href{https://github.com/MAZiqing/FEDformer}{https://github.com/MAZiqing/FEDformer}.
    \item \textbf{Autoformer}:~introduces a decomposition architecture that integrates the series decomposition block as an internal operator, which can progressively aggregate the long-term trend part from intermediate prediction. We implement their provided source code from \href{https://github.com/thuml/Autoformer}{https://github.com/thuml/Autoformer}.
    \item \textbf{iTransformer}:~repurposes the Transformer architecture for time series forecasting by embedding the time points of individual series into variate tokens, enhancing multivariate correlation capture and nonlinear representation learning without altering basic components. We implement their provided source code from: \href{https://github.com/thuml/iTransformer}{https://github.com/thuml/iTransformer}.
    \item \textbf{Crossformer}:~introduces a Transformer-based model that integrates cross-dimension dependencies for MTS forecasting, enhancing temporal and variable interactions via Dimension-Segment-Wise embedding and Two-Stage Attention layer. We implement their provided source code from: \href{https://github.com/Thinklab-SJTU/Crossformer}{https://github.com/Thinklab-SJTU/Crossformer}.
    \item \textbf{WaveForM}: is a graph-enhanced Wavelet learning framework for long-term MTS forecasting. We implement their provided source code from: \href{ https://github.com/alanyoungCN/WaveForM}{ https://github.com/alanyoungCN/WaveForM}.
    \item \textbf{SAN}: proposes a slice-level adaptive normalization scheme for time series forecasting, addressing the challenge of non-stationarity by locally adjusting statistical properties within temporal slices. We implement their provided source code from: \href{https://github.com/icantnamemyself/SAN}{https://github.com/icantnamemyself/SAN}.
    \item \textbf{Dish-TS}: mitigates distribution shift in time series by employing a Dual-CONET framework to separately learn the distributions of input and output spaces, thereby effectively capturing the distribution differences between the two spaces. We implement their provided source code from: \href{https://github.com/weifantt/Dish-TS}{https://github.com/weifantt/Dish-TS}.  
    \item \textbf{RevIN}: is a reversible instance normalization method designed to address the distribution shift problem in time series data by symmetrically removing and restoring statistical information through learnable affine transformations. We implement their provided source code from: \href{https://github.com/ts-kim/RevIN}{https://github.com/ts-kim/RevIN}.
\end{itemize}

\begin{figure}[!t]
\centering
\subfloat[On Electricity]{
    \label{fig8_1}
    \includegraphics[width=1.0\linewidth]{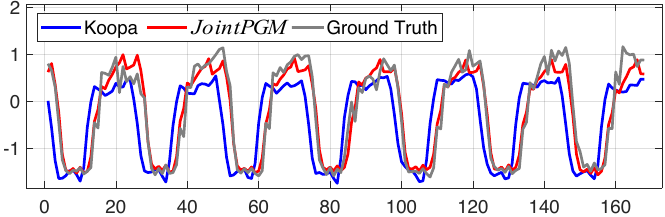}
    }
\hfill
\subfloat[On ETTh1]{
    \label{fig8_2}
    \includegraphics[width=1.0\linewidth]{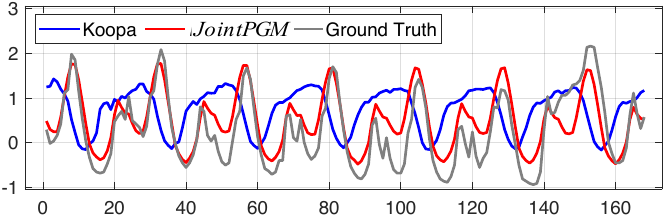}
    }
\hfill
\subfloat[On ETTm2]{
    \label{fig8_3}
    \includegraphics[width=1.0\linewidth]{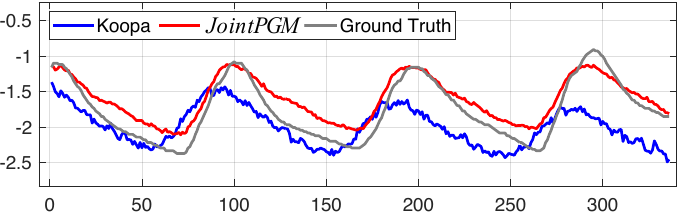}
    }
\hfill
\subfloat[On Exchange]{
    \label{fig8_4}
    \includegraphics[width=1.0\linewidth]{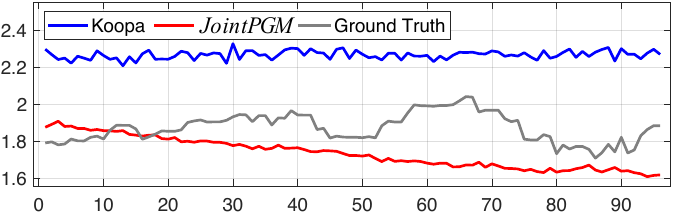}
    }
\hfill
\caption{Visualizing predictions of \textit{JointPGM} and Koopa.}
\label{figure 8}
\end{figure}

\begin{figure}[!t]
\centering
\subfloat[On Electricity]{
    \label{fig9_1}
    \includegraphics[width=1.0\linewidth]{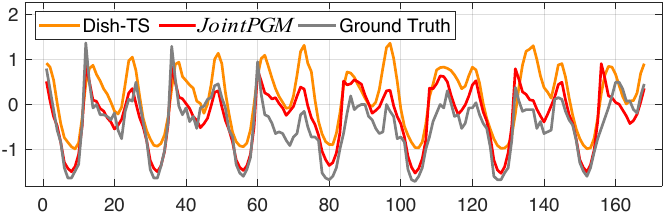}
    }
\hfill
\subfloat[On ETTh1]{
    \label{fig9_2}
    \includegraphics[width=1.0\linewidth]{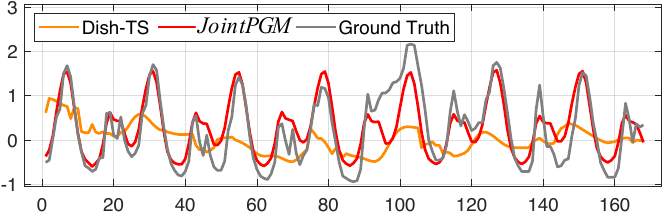}
    }
\hfill
\subfloat[On ETTm2]{
    \label{fig9_3}
    \includegraphics[width=1.0\linewidth]{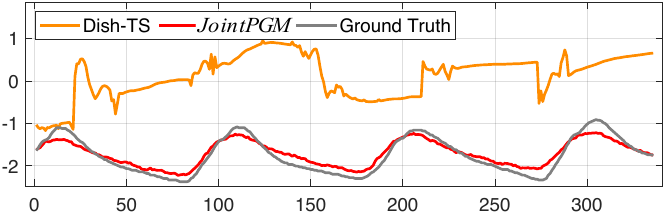}
    }
\hfill
\subfloat[On Exchange]{
    \label{fig9_4}
    \includegraphics[width=1.0\linewidth]{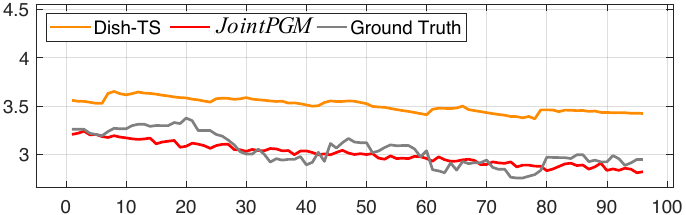}
    }
\hfill
\caption{Visualizing predictions of \textit{JointPGM} and Dish-TS.}
\label{figure 9}
\end{figure}

\begin{figure}[!t]
\centering
\subfloat[On Electricity]{
    \label{fig10_1}
    \includegraphics[width=1.0\linewidth]{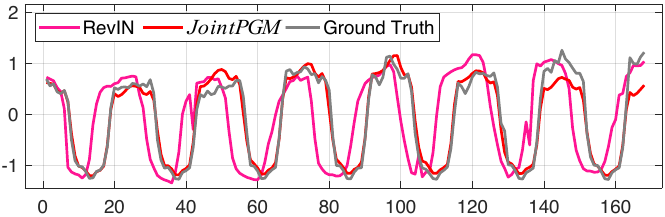}
    }
\hfill
\subfloat[On ETTh1]{
    \label{fig10_2}
    \includegraphics[width=1.0\linewidth]{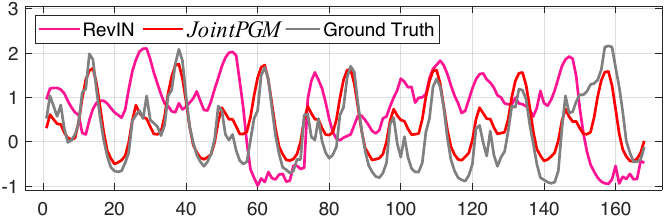}
    }
\hfill
\subfloat[On ETTm2]{
    \label{fig10_3}
    \includegraphics[width=1.0\linewidth]{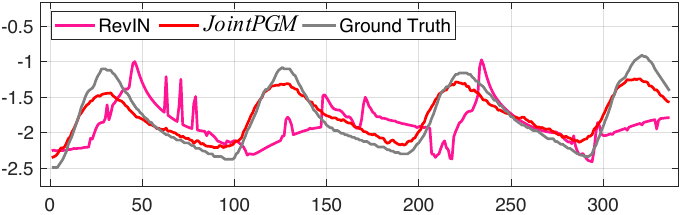}
    }
\hfill
\subfloat[On Exchange]{
    \label{fig10_4}
    \includegraphics[width=1.0\linewidth]{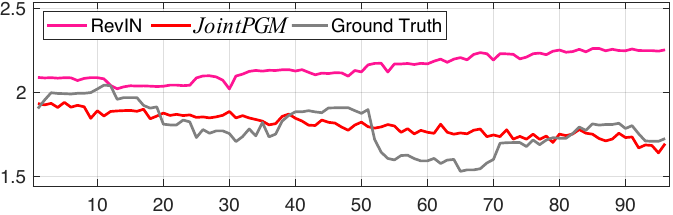}
    }
\hfill
\caption{Visualizing predictions of \textit{JointPGM} and RevIN.}
\label{figure 10}
\end{figure}

\section{More Experimental Results} \label{secd}

\subsection{Comparison with Dish-TS and RevIN} \label{secd.1}
In this section, following the return-to-original-value setting of Dish-TS~\cite{DBLP:conf/aaai/0010WWWZF23}, we further compare the performance with the normalization-based methods Dish-TS~\cite{DBLP:conf/aaai/0010WWWZF23} and RevIN~\cite{DBLP:conf/iclr/KimKTPCC22}, using different backbones including Transformer, Informer, and Autoformer. The results are shown in Table \ref{table C.3}. From the table, it is evident that our \textit{JointPGM} can still achieve SOTA performance in \textbf{75$\%$} of forecasting results compared with Dish-TS and RevIN.

\subsection{Model efficiency} \label{secd.2}
Figure \ref{figure 5} shows the supplementary efficiency results for ETTm2 (7 variables, 69680 time steps).
We observe that in datasets with relatively few variables and a large time step scale (ETTm2), the efficiency of \textit{JointPGM} is only slightly inferior to DLinear, PatchTST, and iTransformer.
For example, when compared to the MLP-based model Koopa customized for non-stationary forecasting, \textit{JointPGM} achieves a reduction of 58.9$\%$ in training time for the ETTm2 dataset, while maintaining a memory footprint of only 97.8$\%$.

\subsection{Heatmap visualization of inter-series correlation} \label{secd.4}
To further provide an intuitive understanding of the learning processes of the stacked intra-series learner and inter-series learner, we present some inter-series correlation visualizations on the METR-LA and Electricity test set in Figure \ref{figure 13}. Concretely, we calculate the Pearson Correlation coefficients for each pair of series in $\bm{X}_{t-L:t}$, $\hat{\bm{H}}_{t-L:t}$, $\tilde{\bm{H}}_{t-L:t}$ and $\bm{X}_{t:t+H}$, and visualize the entire correlation matrix. 
It can be clearly observed that there is an obvious variation in inter-series correlation between lookback window $\bm{X}_{t-L:t}$ and horizon window $\bm{X}_{t:t+H}$, reflecting a significant shift in distribution between these two windows.
As we observe in the shallow intra-series learner, we find that the correlation of representation $\hat{\bm{H}}_{t-L:t}$ is similar to the correlation of the lookback window $\bm{X}_{t-L:t}$. As we go deeper into the inter-series learner, the correlation of representation $\tilde{\bm{H}}_{t-L:t}$ gradually becomes more similar to the correlation of the horizon window to be predicted.
This observation verifies that our proposed \textit{JointPGM} effectively addresses the transitional shift between the lookback and horizon windows.

\subsection{Visualization of forecasting results} \label{secd.3}
To offer a clear comparison between various models, we show supplementary forecasting showcases on Electricity, ETTh1, ETTm2, and Exchange datasets: Figure \ref{figure 8}, Figure \ref{figure 9} and Figure \ref{figure 10} show the predictions of our \textit{JointPGM} and three baselines Koopa, Dish-TS and RevIN respectively.
We can see that when the series trend changes dramatically, our \textit{JointPGM} can still acquire accurate predictions. These visualizations illustrate the effective forecasting capability of \textit{JointPGM} in handling shifted multivariate time series.

\end{document}